\documentclass[11pt, a4paper, logo, twocolumn, copyright]{googledeepmind}

\pdftrailerid{redacted}

\makeatletter
\renewcommand\bibentry[1]{\nocite{#1}{\frenchspacing\@nameuse{BR@r@#1\@extra@b@citeb}}}
\makeatother

\usepackage{dsfont}

\usepackage[authoryear, sort&compress, round]{natbib}

\usepackage[font=small,labelfont=bf]{caption}
\usepackage{multirow}

\graphicspath{{figures/}}

\title{GATS: Gather-Attend-Scatter}

\correspondingauthor{kondiz@google.com}

\renewcommand{\today}{}

\author[ \hspace{-0.08cm}]{Konrad~\.Zo\l{}na}
\author[ \hspace{-0.08cm}]{Serkan~Cabi}
\author[ \hspace{-0.08cm}]{Yutian~Chen}
\author[ \hspace{-0.08cm}]{Eric~Lau}
\author[ \hspace{-0.08cm}]{Claudio~Fantacci}
\author[ \hspace{-0.08cm}]{Jurgis~Pasukonis}
\author[ \hspace{-0.08cm}]{Jost~Tobias~Springenberg}
\author[ \hspace{-0.08cm}]{Sergio~G\'{o}mez~Colmenarejo}

\affil[ ]{All authors are affiliated with Google DeepMind}

\begin{abstract}
As the AI community increasingly adopts large-scale models, it is crucial to develop general and flexible tools to integrate them. 
We introduce Gather-Attend-Scatter (GATS), a novel module that enables seamless combination of pretrained foundation models, both trainable and frozen, into larger multimodal networks.
GATS empowers AI systems to process and generate information across multiple modalities at different rates.
In contrast to traditional fine-tuning, GATS allows for the original component models to remain frozen, avoiding the risk of them losing important knowledge acquired during the pretraining phase.
We demonstrate the utility and versatility of GATS with a few experiments across games, robotics, and multimodal input-output systems.
\end{abstract}

\begin{document}

\maketitle

\section{Introduction}

Our world is inherently multimodal, with information encoded in a variety of modalities such as text, images, and video.
Understanding and interacting with the world effectively requires the ability to process and reason across these modalities, keeping in mind that information from different modalities often arrives at different points in time and at different rates.

Advances in deep learning have led to significant progress in unimodal or bimodal tasks, but the development of multimodal models remains a challenge.
One of the key obstacles is the scarcity of high-quality, aligned, multimodal data.

\begin{figure}[t]
	\centering
	\includegraphics[width=0.95\columnwidth]{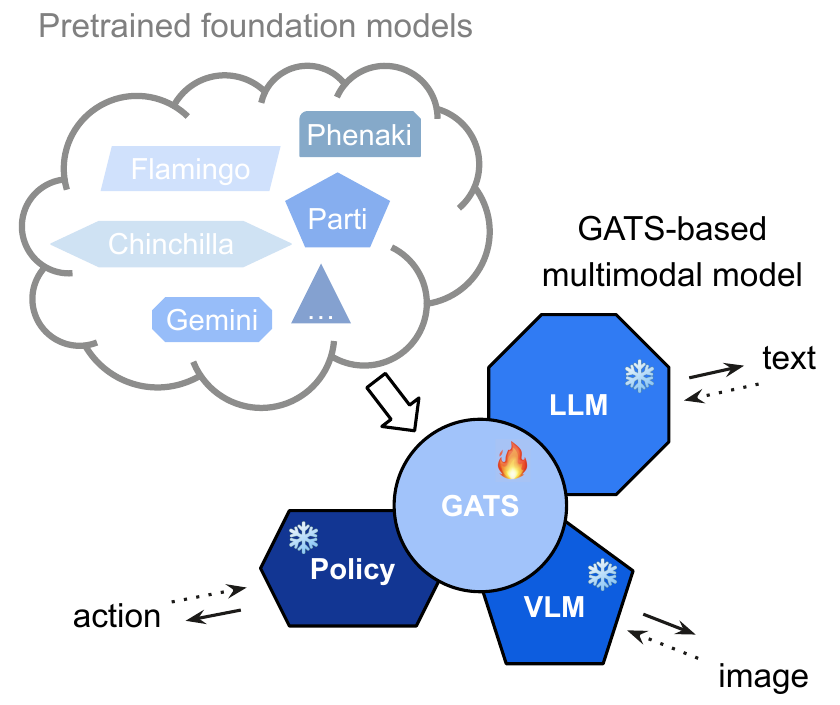}
	\caption{
	By leveraging GATS, a modular multimodal architecture can be constructed, integrating frozen pretrained foundation models.
	\label{fig:gats_intro}}
\end{figure}

To make progress in this area, we propose GATS (Gather-Attend-Scatter), a module that enables us to combine and re-use a wide range of pretrained models (e.g. vision, language, proprioception, actions) to produce models with multimodal inputs and outputs. Moreover, the resulting architecture can process information from different modalities at different rates, making it well suited to robotics.
Figure~\ref{fig:gats_intro} illustrates the GATS approach.

GATS works by gathering activations from all component models, attending to the most relevant information, and scattering the combined representation back to all models by modifying their original activations.
Hence, instead of simply combining representations from different models, GATS leverages a property of neural networks shown by \citet{Flamingo} that a given network's behavior can be reprogrammed by modifying its activations.

The proposed approach is very general and can be applied to any deep neural network, as they all process their activations layer by layer.
As such, GATS is agnostic to the specific details of the neural networks being combined.
The resulting GATS multimodal architectures only require the GATS module to be trained, thus eliminating the need to finetune the original pretrained models and preventing potential loss of knowledge.
This makes GATS a highly versatile and general-purpose tool for building multimodal models from a variety of pretrained sources.
%

% The contributions of this work are as follows,
% \begin{enumerate}
%     \item We introduce a new general light-weight module that enables communication among foundation models, each of which can be pretrained or co-trained. The new module generalizes the interleaving cross-attention layer in \citep{Flamingo} to combine an arbitrary number of network sources in a unified style.
%     \item We train an efficient robotic agent based on the GATS module and utilize large scale text and vision foundation models pretrained separately on general domain datasets.
%     \item We also demonstrate GATS' general applicability in other domains with a bimodal network that can be used as both image captioner and text-to-image generator.
% \end{enumerate}

\section{Gather-Attend-Scatter in detail}
In this section, we describe the GATS module in detail.
Specifically, we explain how GATS enables communication between multiple pretrained neural networks.
While GATS can be used in tandem with any deep learning model, for the sake of clarity, we will focus on its application to transformer-based component models.

The proposed architecture of the GATS module consists of several GATS layers similar to vanilla transformer layers equipped with local attention.
GATS layers are interleaved with all given networks and hence serve as a bridge between them.
While the component networks process only tokens of their native modality, GATS attends to embeddings of all networks and thus modalities.

\subsection{GATS layer}\label{sec:gats_layer}
Before explaining how GATS interacts with component networks, we describe a single GATS layer, its input and its architecture.

\begin{figure}[t]
	\centering
	\includegraphics[width=0.95\columnwidth]{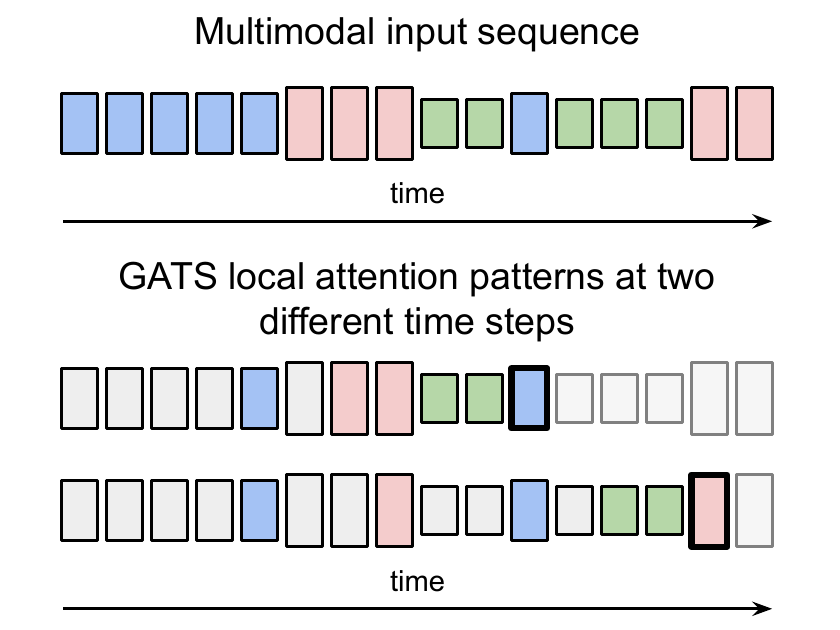}
	\caption{
	\textbf{Top.}~GATS can process sequences of activations from different models, and hence with different sizes.
	Each element's color corresponds to its modality.
	The same-colored embeddings have the same size.
	\textbf{Bottom.}~Two examples.
	The GATS module has separate local context lengths for each modality.
	The visualisations show which embeddings are visible to the GATS module when the bold embedding is processed, assuming that each modality has the same context length of 2 (grayed embeddings are ignored by GATS and do not take part in its computations).
	\label{fig:gats_layer}}
\end{figure}

\paragraph{Input}
We assume that the input to a GATS layer is a sequence of embeddings $(x_1, x_2, \dots, x_t)$ and each element belongs to one of $M$ modalities, i.e., there is a function $m(x_i)$ such that $m(x_i) \in \{1, 2, \dots, M\}, \forall_{i}$.
In our case each $x_i$ is an input embedding or an activation of a neural network.
Importantly, we do \emph{not} assume that sizes of $x_i$ and $x_j$ are the same, as in practical situations the activations of pretrained models almost always vary in size. 
We do, however, assume that the sizes are the same if $x_i$ and $x_j$ belong to the same modality, i.e., if $m(x_i) = m(x_j)$.
See Figure~\ref{fig:gats_layer} (Top) for a visualization of the input structure.

\paragraph{Gather}
The most notable difference between GATS and a vanilla transformer is the gather step.
While conventional transformers attend to all the recent embeddings that fit in their context length $N$, GATS employs a special local attention and operates on the last few embeddings from each modality.
Specifically, its context length is split over all modalities $N = N_1 + N_2 + \dots + N_M$, where $N_m$ is the context length for the $m$-th modality.
This important choice makes GATS attend to embeddings from all modalities processed so far, even if recent inputs come from only a single modality.
The memory lengths $N_m$ can be relatively small, assuming that long term processing is delegated to the pretrained models.
See Figure~\ref{fig:gats_layer} (Bottom) for visualizations.

In precise mathematical terms, for the input sequence of embeddings $(x_1, x_2, \dots, x_t)$, the gather step takes the largest subsequence $G$ such that
\begin{itemize}
    \item $\lvert \{ x_i : x_i \in G \land m(x_i) = m \} \rvert \leq N_m$, i.e., the number of elements of G belonging to the modality $m$ is no larger than $N_m$\footnote{The number of elements is lower than $N_m$ only if there are not enough elements from the modality $m$ in the input.},
    \item $\forall_{i<j} \bigl(x_i \in G \land m(x_i) = m(x_j) \bigr) \implies x_j \in G$, i.e., the most recent embeddings are in $G$.
\end{itemize}
Only the selected subsequence $G$ takes part in the next GATS processing, i.e., $x_i \notin G, \forall i \in \{1, \dots,\ t \}$ are ignored.

\paragraph{Attend}
The next step is an attention over the gathered subsequence $G$.
GATS deals with different sizes of the embeddings in $G$ in the following manner: for each modality $m$, there is a projection $p_m$ such that the size of $p_m(x_i)$ is $d$.
Importantly, $d$ is relatively small and the same for all projections.
Each embedding from $G$ is projected with a corresponding projection
\begin{align}
    x_i \leftarrow p_{m(x_i)}(x_i),
\end{align}
which results in obtaining a new sequence where all elements have the same dimensionality $d$.

Once the subsequence G is projected, it is further processed by a vanilla transformer layer (i.e., single self-attention $Attention$ and single feed-forward layer $FFW$), i.e., for each $x_i \in G$ we have
\begin{align}
    x_i \leftarrow FFW(Attention(p_{m(x_i)}(x_i), G)).
\end{align}
We also apply the usual layer norms, residual connections, and positional encodings in our experiments, but for the sake of clarity, we assume they are folded inside $Attention$ and $FFW$ functions, and hence do not have to be explicitly written in the equation.
Please see the original transformer paper~ \citep{vaswani2017attention} and the supplementary materials for details.

\paragraph{Scatter}
The processed embeddings are then "projected back", i.e., for each modality $m$ there is a function $r_m$ such that the sizes of the initially input $x_i$ and the final $r_m(FFW(Attention(p_{m(x_i)}(x_i), G)))$ are the same.
This guarantees that the sizes of the gathered subsequence $G$ and the resulting output sequence~match.

Before the projected embeddings are finally output, a gated residual connection is added.
A value of a gate is a scalar in $[0, 1]$ computed by a gating function $g_m$ that has the same inputs as~$r_m$.

To summarize all GATS layer transformations, each selected $x_i \in G$ is processed in the following~way
\begin{align}
    x_i &\leftarrow x_i + g_m(z_i) \cdot r_m(z_i),
\end{align}
where $z_i = FFW(Attention(p_{m(x_i)}(x_i), G))$ is the output of the attention block.
Unselected input elements ($x_i \notin G$) stay unaltered.

\subsection{Interleaving with vanilla transformers}\label{sec:gats_interleaving}
A GATS module could in theory be trained from scratch on multimodal data.
However, as mentioned before, we focus our study on the use case of GATS interleaved with pretrained transformer-based component models.

While the pretrained networks process only embeddings from their native modalities, GATS sees all the embeddings put together (but processes only a few of them due to its gather step, as described in Section~\ref{sec:gats_layer}).
This important property makes GATS capable of being interleaved with all the component networks.
As a result, each of the component models is conditioned on processing done so far by all other models.
This section describes the details of the interleaving process.

For simplicity, we assume in this section that all pretrained transformers are unimodal, but GATS readily accommodates cases where inherently multimodal models are fused with those operating on different modalities.

\paragraph{Interleaving arrangement}
Since the number of layers of given transformers vary, interleaving them with GATS may at first seem complex.
However, we see that simple proportional interleaving works well in practice.
Specifically, if GATS has $K$ layers, and the given transformers have $L_1, L_2, \dots, L_M$ layers, inputs to the $k$th GATS layer are embeddings obtained from $l_{k,1}, l_{k,2}, \dots, l_{k,M}$ layers (respectively for each model) where
\begin{align}
    l_{k,i} = \min(\max(1, \lfloor k \frac{L_i}{K} \rfloor), L_i - 1).
\end{align}
We say that the $k$th layer of GATS lies between $l_{k,m}$th and $l_{k,m}+1$th layers of a unimodal transformer $m$.
The $\min$ and $\max$ are used to guarantee that the value of each $l_{k,i}$ is positive and never larger than $L_i - 1$.

\begin{figure}[t!]
	\includegraphics[width=0.95\columnwidth]{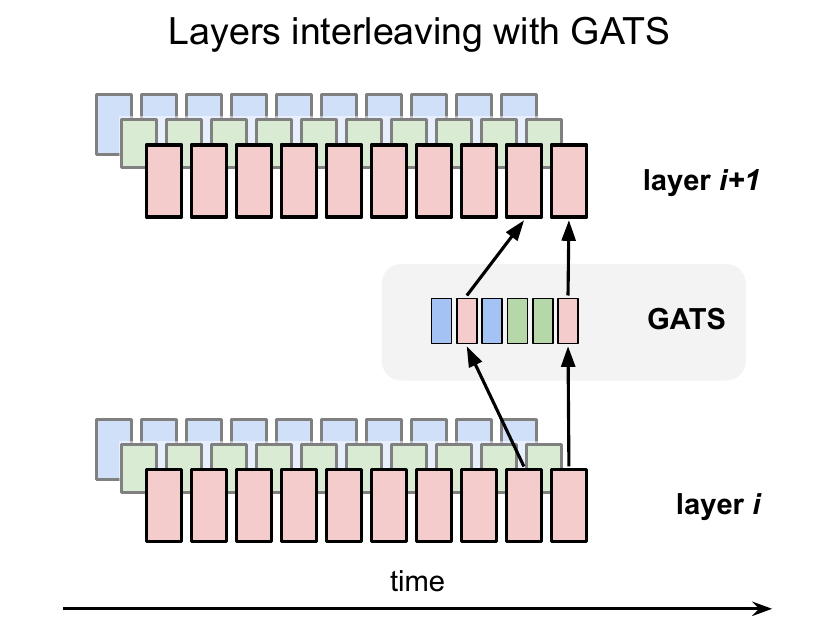}
	\caption{
	GATS interleaves with the red unimodal transformer processesing only red embeddings.
	GATS gathers two recent embeddings from each unimodal network, projects them to the common size, attends over them and scatters the output.
	The next layer of the red transformer processes activations altered by GATS instead of the original ones.
	\label{fig:gats_interleaving}}
\end{figure}

\paragraph{GATS interleaving}
If a given layer of GATS lies between the $i$th and $i+1$th layers of a unimodal transformer, the output of the $i$th layer, instead of being fed immediately to the $i+1$th layer, is first processed and updated by the GATS layer.
Since the same GATS layer is also interleaved with other networks, currently processed embeddings interact with past embeddings from other modalities.
That enables information flow between models.
See Figure~\ref{fig:gats_interleaving} for a visualization.

\paragraph{Steering}
While a GATS layer updates all input activations, it is not clear if all component models should use updated activations in the following layer.
For example, one may want to only condition the multimodal model via GATS on activations from a large-scale pretrained language model while leaving the language model processing itself out from conditioning on other modalities.
If a component model uses updated activations, we say that GATS \emph{steers} it.
Note that for a modality $m$ that is not steered, the function $r_m$ is not used (and hence does not have to be trained), and corresponding embeddings (i.e., $x_i$ such that $m(x_i) = m$) can be used only to build key and values in GATS attention and are skipped from being queried to further save computation.
Currently, we manually specify which modalities are steered, but this could potentially be learned from data.

\subsection{Hyperparameters}
A GATS layer is defined by the following hyperparameters:
\begin{itemize}
    \item number of given unimodal models $M$,
    \item local context lengths $N_1, N_2, \dots, N_M$,
    \item projected embedding size $d$,
    \item a nonempty subset $S$ of $\{1, 2, \dots, M\}$ specifying which component models are steered,
    \item hyperparameters describing the transformer used to process embeddings in the local context $G$.
\end{itemize}

A GATS module consists of $K$ GATS layers.
In our experiments all layers share the same hyperparameters but in theory all of them can be different.
We use simple linear transformations ($2M$ of them) for all $p_m$ and $r_m$ functions.
Each of the gating functions $g_m$ is also a linear transformation additionally followed by a layer norm.
As such all $p_m$, $r_m$ and $g_m$ functions depend only on the projected embedding size $d$ (and input embeddings sizes which are usually out of our control) and hence do not require additional hyperparameters.
Finally, an alternative interleaving arrangement could be used but we always use the proportional one described in Section~\ref{sec:gats_interleaving}.

\subsection{Example 1: cross-attention via GATS}\label{sec:gats_flamingo_example}
To showcase GATS generality and help build an intuition around it, we present a very special and simple case where a pretrained language model is conditioned on visual features to caption images.
The resulting architecture is very similar to a vision-to-text cross-attention model presented in Flamingo by \citet{Flamingo}, a state-of-the-art method for this task.

\begin{figure}[t]
	\centering
	\includegraphics[width=0.95\columnwidth]{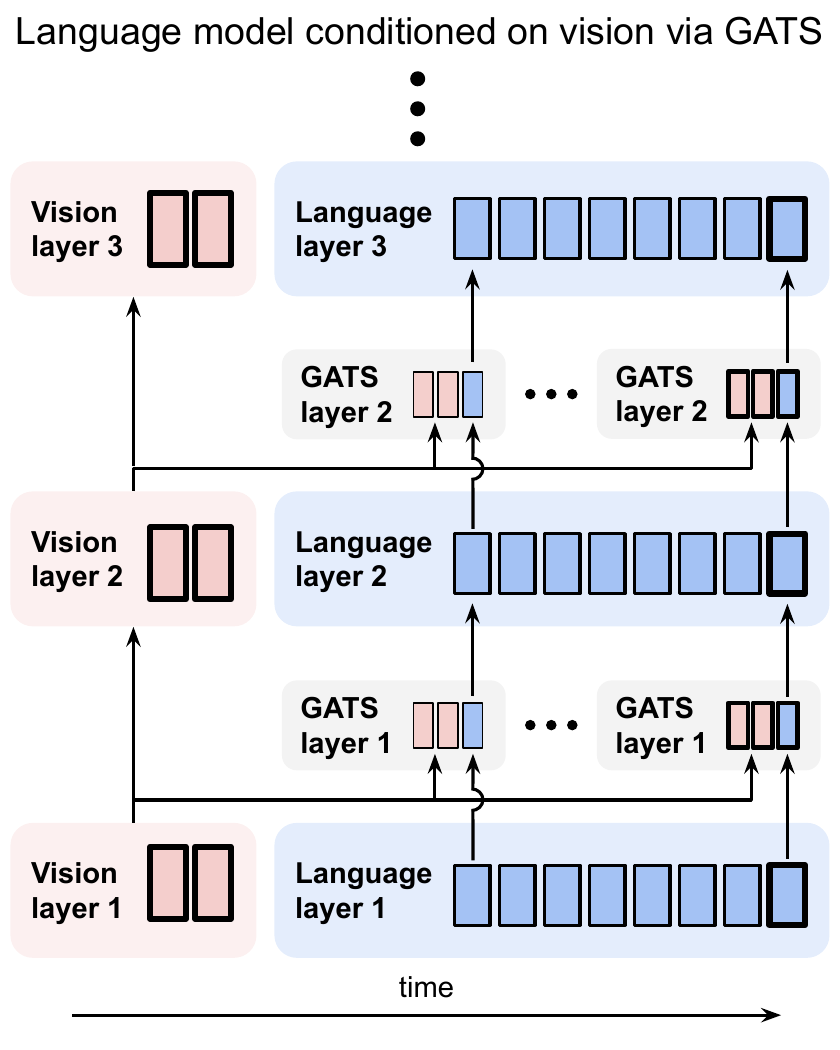}
	\caption{
	A special case of a GATS-based architecture that acts as a typical vision-to-text cross-attention model.
	Vision features (two red embeddings) are obtained in the beginning and are always visible to GATS, as its vision context is set to two.
	GATS language context is one and hence only the most recent language token is gathered.
	GATS steers the language model by updating the language activations and effectively conditions language processing on vision.
	Bold tokens indicate those processed (gathered, attended or scattered) by GATS when the most recent language token is processed.
	The figure shows an example with a single image, but interleaving text and images is a straightforward extension.
	\label{fig:gats_flamingo_example}}
\end{figure}

We assume that there are two pretrained models given: a vision model and a large language model (i.e., there are two models $M=2$, and we set $1$ for the vision model and $2$ for the language model).
For a given image, the vision model is run first to obtain $V$ visual features.
This is followed by the frozen language model interleaved with GATS.
GATS steers only the language model (vision processing is not altered, i.e., we have $S=\{2\}$), and the GATS context lengths are set to always cover all vision features (i.e., $N_1 = V$) and the last text embedding (i.e., $N_2 = 1$).
The projected embedding size $d$ is set to match the language transformer activation size, and hence $p_2$ and $r_2$ transformations can both be identity functions.
Additionally, to resemble the Flamingo architecture even more, extra text embeddings representing the image position within the text can be added.
See Figure~\ref{fig:gats_flamingo_example} for a visualization.

\subsection{Example 2: GATS-based robotic agent}\label{sec:gats_agent_example}

In this section we show how we use the GATS module to combine a pretrained language model and a pretrained video model into a single model driving a robot capable of instruction following.
Hence, we have three modalities.
\begin{enumerate}
    \item Language: an instruction specifying the desired behavior of the robot, given at the beginning of each episode.
    \item Video: frames representing visual input to the robot, given frame by frame at each environment time step.
    \item Action: proprioception values (including model outputs i.e., values comforting action specifications). Similarly to video frames, given at each environment time step.
\end{enumerate}
Since there is an abundance of unpaired language and video data, and also corresponding large-scale pretrained models, we keep these generalist models frozen and extract good features through GATS steering.
In contrast to this, we train action model from scratch to predict discretized action tokens, as it is strongly coupled with the robot setup.
The GATS module is also trained from~scratch.

During an episode, the language model is called only once to generate activations for a given instruction.
These activations can be cached for frequent instructions, enabling the use of a larger language model without sacrificing inference speed.
The size of video and action models should be tuned such that the desired inference speed is achieved.

\begin{figure*}[hp!]
	\centering
	\includegraphics[width=1.9\columnwidth]{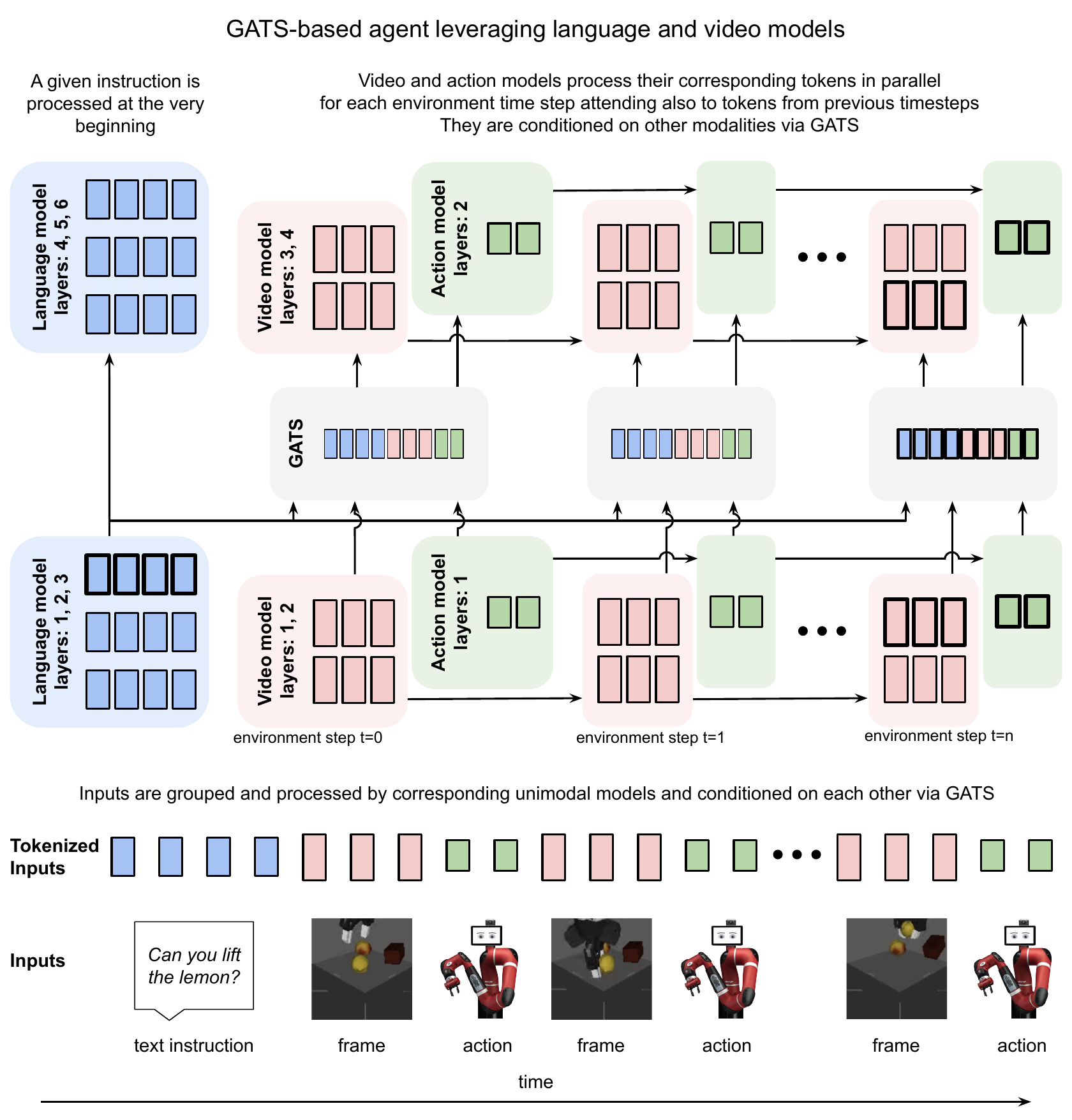}
	\caption{
    This figure showcases a GATS-based agent architecture leveraging pretrained language and video models.
    For clarity, we use small models with 6, 4, and 2 layers for language, video, and action respectively, and a single layer for the GATS model.
    Colors represent different modality, blue for language, red for vision and green for action.
    The workflow begins with a single text instruction processed once (the step has to be repeated if another instruction is given).
    Then, with each environment step, a video frame and proprioception inputs are fed into their respective unimodal models and are processed alongside previous tokens from the same modality (i.e., the unimodal models have contexts long enough to fit more than a single environment step).
    GATS gathers activations from the language model's 3rd layer, the video model's 2nd layer, and the action model's 1st layer, using them to steer further processing for the two latter models.
    Importantly, GATS only attends to recent activations, delegating long-term processing to the individual models, resulting in negligible computational overhead.
    This design allows for seamless scaling of both model size and GATS layers, making the architecture highly flexible and adaptable.
    Bold tokens indicate those processed (gathered, attended or scattered) by GATS for the most recent environment step, highlighting the efficient interaction between GATS and the unimodal models.
	\label{fig:gats_agent_example}}
\end{figure*}

For each environment step, a video frame gets tokenized and processed by the video model.
Simultaneously, the action model processes proprioception inputs.
Importantly, these models are equipped with long context and condition their outputs also on the tokens from the previous time steps.
Both models are interleaved with GATS modules that process video and action activations from the most recently processed environment step alongside precomputed language activations.
GATS outputs are used to steer the video and action models.
Please see Figure~\ref{fig:gats_agent_example} for a visualization of the entire system.

Adding new modalities such as audio would be straightforward.
They could even be added to a system already trained on a subset of modalities.
In that case, GATS' parameters relating to the new modality require initialization from scratch but other parameters could be seamlessly transferred from the previous iteration.
Even an abstract modality, uncoupled from real-world data such as a "scratch pad" for extra computations, could be added.
Finally, a single modality can be processed by multiple models, like a fast low-resolution video model for dynamics alongside a larger image model for manipulation details that runs periodically.
Replacing a unimodal model with another one is just as simple.

One major drawback of monolithic architectures like Gato~\citep{reed2022} is their sequential token processing, forcing modalities to wait for each other.
The monolithic models also allocate equal compute to each token, making leveraging truly large-scale models (e.g., for language) impractical.
GATS-based architectures overcome this problem, as tokens from each modality can be processed by the corresponding unimodal models simultaneously.
For example, let us consider how easy it would be to add a "talking" capability to our system presented in Figure~\ref{fig:gats_agent_example}: simply output text tokens when the robot needs to speak.
Text tokens would be occasionally generated by the language model conditioned on other modalities via GATS.
Importantly, this extra language processing could be fully decoupled and operate on its own rate, as it would be orthogonal to the ongoing video and action processing stream, so the robot behavior would not be disturbed.

\subsection{Properties}
In this section we highlight properties of GATS that make it suitable as a universal connector for pretrained models.

\paragraph{Generality}
GATS is symmetrical in a sense that no modality has to be treated in any special way.
Each component model processes its own inputs and resulting activations may be used to aid all other models.
GATS is exposed to these activations taken from different layers, and is trained on how to process them (as opposed to hard-coding which layers are used, e.g., middle versus ultimate ones).
That being said, through a modest set of hyperparameters, GATS can be set up to act as highly asymmetrical specialist architectures such as vision-to-text cross-attention model (see details of this example in Section~\ref{sec:gats_flamingo_example}).

\paragraph{Lightweight inference overhead}
A given input embedding is processed by only one corresponding unimodal model, and hence the inference speed for a considered modality does not depend on sizes of other component models.
The processing is, however, conditioned on the other modalities thanks to GATS steering.
Importantly, only already processed activations from other modalities are gathered by GATS.
This means that the only added overhead is caused by running GATS layers, but this is relatively lightweight, as the size of GATS projected embeddings $d$ are small and GATS local context lengths $N_1, N_2, \dots, N_M$ are short.

Additionally, to further limit extra computations, the number of GATS layers can be significantly smaller than the numbers of layers in component models.
It means that modalities paired with smaller models would be still fast to process, while exposed via GATS to rich representations from larger models (e.g., large-scale pretrained language model).

\paragraph{Efficient training}
The entire multimodal model consisting of all component transformers and the GATS module can be trained very efficiently, since GATS can be parallelized the same way vanilla transformers are.
Additionally, thanks to steering, GATS accesses information stored in weights of frozen pretrained models without the need to spend device memory on their updates.

\section{Model components}

In this section we describe a simple way to adapt classifier-free guidance \citep{ho2022classifier} to improve language conditioned agents.
We also list and discuss all the pretrained foundation models used in our experiments.

\subsection{Classifier-free guidance}\label{sec:cfg}

Classifier-free guidance has been widely adopted in diffusion-based generative models to improve the output alignment with the conditioning input.
The same idea has also been applied to agent training \citep{lifshitz2023steve}.

Let $l(x,c)$ and $l(x)$ be the logit vector of the predicted agent action before softmax given observation $x$, with or without language conditioning $c$ respectively.
The guided policy is computed as
\begin{align*}
    \pi(x, c) = \mathrm{Softmax}\left(l(x,c) +
    \lambda (l(x,c) - l(x)) \right),
\end{align*}
where scalar $\lambda \geq 0$ controls the guidance strength.
When $\lambda = 0$, it is the standard conditional generative model.
When $\lambda > 0$, it puts higher probability on tokens where they are more likely with the given conditioning input compared to the marginal distribution.
In our experiments we set $\lambda=0.5$.
To obtain an agent network we compute both conditional and unconditional policies.
We randomly mask the text input with a probability of 0.02 during training.
Enabling the classifier-free guidance consistently improves the performance of our agents (see ablation experiments in Section~\ref{sec:agent_exp_lan_table}).

\subsection{Pretrained models}\label{sec:pretrained_models}

To demonstrate GATS's generality as a tool for unifying diverse foundation models, we conducted experiments employing a variety of pretrained models trained on rich data across text, image, and video modalities.

\paragraph{Chinchilla}
We used pretrained language models from the Chinchilla family \citep{hoffmann2022training}.
While the original family includes models ranging from 70 million to over 16 billion parameters, we experimented mainly with the 1.3 billion instance.
The model uses a coupled text tokenizer that parses input strings to get text tokens that are further processed by a transformer.

\paragraph{Phenaki}
We used the Phenaki video model \citep{villegas2022phenaki} with 1.1B parameters to process video inputs with text conditioning.
It first uses a C-ViViT encoder to convert $2T+1$ 128x128 frames of a video sequence into $T+1$ 16x16 discrete token frames, then applies random masking, and passes through a transformer to predict the masked tokens.
To process text conditioning inputs, it uses the T5 tokenizer \citep{raffel2020exploring} and trains another transformer jointly to encode text and feed into the main video transformer stack.
In our experiments, we disable masking and use all the video tokens to extract features.
We also reuse the Phenaki text encoder as our text module for the GATS-based video agent.

\paragraph{Parti}
While we do not use the entire Parti model proposed by \citet{yu2022scaling}, we used their image tokenizer.
The tokenizer operates on images of resolution 256x256 and was used in all experiments where an image model (as opposed to a video model) is a vision core.

\paragraph{ViT}
We pretrained our own vision transformers \citet{dosovitskiy2021image} on acquired image-text data.
The vision transformer architecture was identical to Chinchilla language models used in our experiments~\citep{hoffmann2022training} and was conditioned on a 1.3 billion Chinchilla language model via GATS.
We trained two variants sized 2.7 billion and 8.5 billion parameters in total (i.e., including 1.3 billion parameters for the frozen language model).
See Section~\ref{sec:text_and_image} for details.

\section{Agent experiments}\label{sec:agent_exp}

We first evaluate a GATS-based agent on a controlled task, a simple Atari game, utilizing only vision and action modalities.

This provides a foundation for subsequent experiments in the more complex Language-Table environment implementing three modalities: language, vision, and action.
We use this open-sourced environment to conduct comprehensive experiments and perform ablative analyses to elucidate the inner workings of GATS and reveal its strengths.

Next, we move to the YCB environment, a more challenging domain that demands manipulation of objects given multiple camera viewpoints.
While sharing the same trimodal structure as Language-Table, YCB necessitates greater precision and dexterity.

\subsection{Environments and Datasets}

This section outlines the environments used, obtained datasets, and evaluation setup.

\paragraph{Atari Pong}
We use the Atari Pong game from the Arcade Learning Environment \citep{bellemare2013arcade} for our evaluations and train on the data used in \citep{reed2022}.

The goal of the game is to outmaneuver the opponent (built-in bot) by controlling a paddle and deflecting a ball across a divided screen, preventing it from crossing your side and scoring points.
The game ends if the player or the bot achieves~21 points.

\begin{figure}[t!]
	\centering
	\includegraphics[width=1.0\columnwidth]{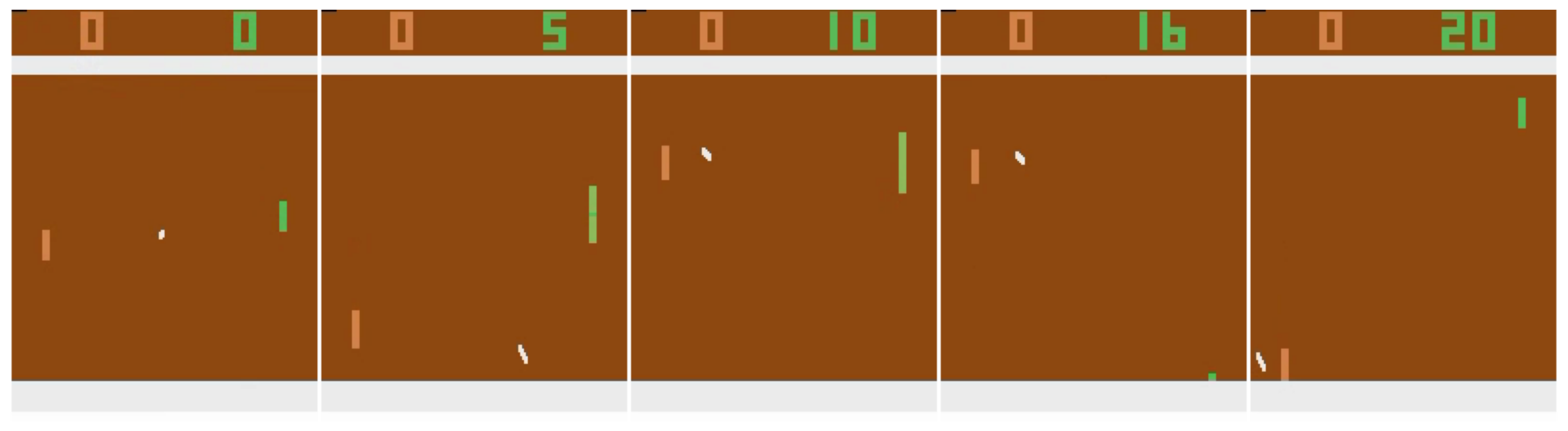}
	\includegraphics[width=1.0\columnwidth]{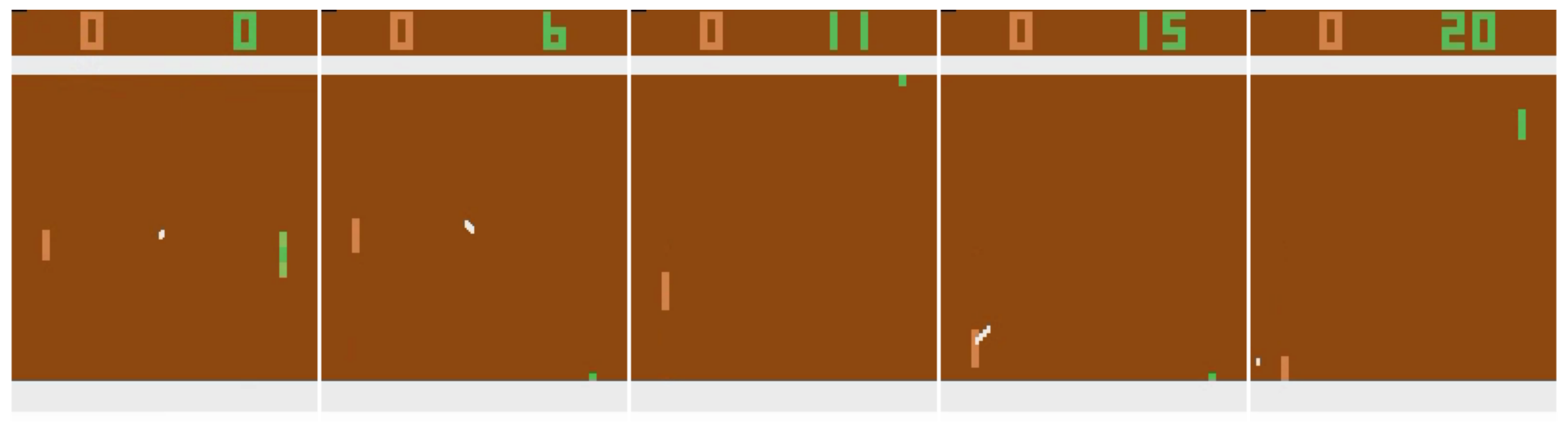}
	\includegraphics[width=1.0\columnwidth]{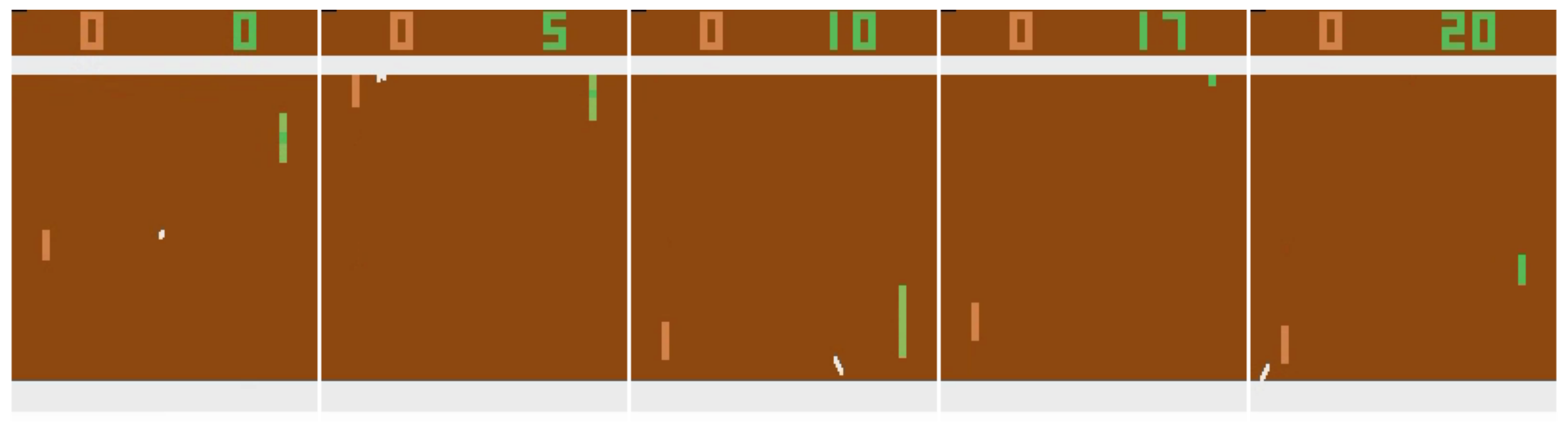}
	\caption{Three episodes in the Atari Pong environment.
	We took 5 snapshots from each episode. The leftmost images show initializations, and the rightmost ones are the final frames.
	Our agent wins both games without losing a single point.
	\label{fig:atari_episode}}
\end{figure}

Agents only have access to the game screen (see Figure~\ref{fig:atari_episode}) and joystick actions are coded as a single $[0, 17]$ integer.

The data consist of $20,000$ episodes sampled randomly across a Muesli \citep{hessel2021muesli} RL training run.
The average Pong reward of the fully trained Muesli agent is 17.89.

\paragraph{Language-Table}
We use the open-source Language-Table benchmark \citep{lynch2022interactive}.
This simulated environment presents a tabletop manipulation scenario, which consists of a robotic arm with a cylindrical end-effector constrained to move in a 2D plane, and a set of 8 blocks with different shapes and colors.
The goal is different for every episode and is presented as a language instruction (e.g., "push the blue triangle to the top left corner").

All of our evaluations are run on the open-source simulated environment using the canonical five task families described in \citep{lynch2022interactive}.
We report the average success rate obtained across 500 episodes (100 per family).

\begin{figure*}[hbt!]
	\centering \footnotesize 
	\includegraphics[width=2.0\columnwidth]{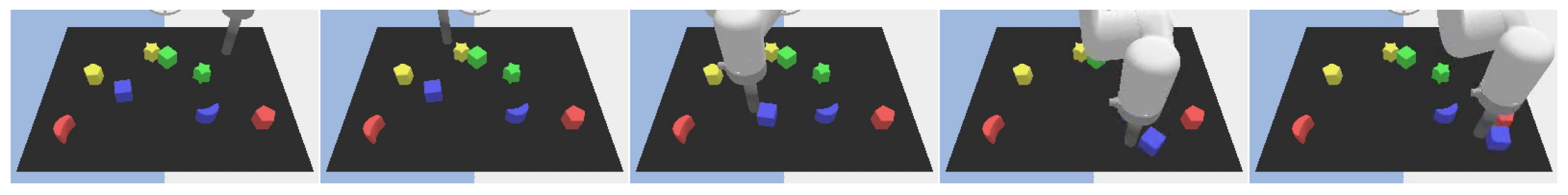}
	\textit{Instruction: move the blue cube to the lower right corner.}
	\includegraphics[width=2.0\columnwidth]{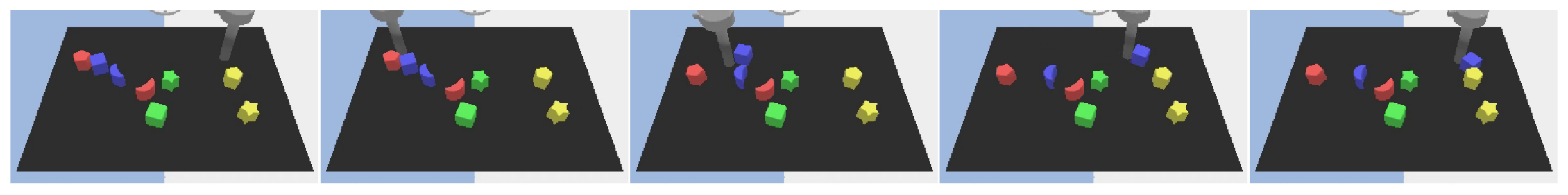}
	\textit{Instruction: push the blue cube close to the yellow pentagon.}
	\includegraphics[width=2.0\columnwidth]{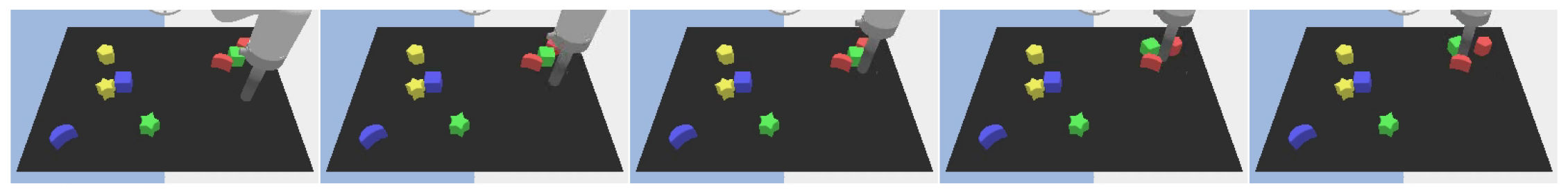}
	\textit{Instruction: separate the red moon from the green cube.}
	\caption{
    Three episodes in Language-Table environment and corresponding instructions given to the agent.
    We took 5 snapshots from each episode. The leftmost images show initializations, and the rightmost ones are the final frames.
	\label{fig:lan_table_episode}}
\end{figure*}

We use a single top-down camera view as observation  (see Figure~\ref{fig:lan_table_episode}).
Actions are discretized 2D delta Cartesian setpoints.

In our experiments we use open-sourced data\footnote{\url{github.com/google-research/language-table}}, a mix of $442{,}226$ real episodes and $181{,}020$ simulated episodes collected at 5hz.
We re-weight each batch such that 75\% of the trajectories originate from simulated episodes.

\paragraph{YCB}
This simulated environment is written with Google DeepMind’s open-source MoMa library\footnote{\url{github.com/google-deepmind/dm_robotics/blob/main/py/moma/README.md}}.
MoMa uses MuJoCo~\citep{todorov2012mujoco} underneath as
a physics engine for the simulation.

\begin{figure}[t]
	\centering
	\includegraphics[width=0.95\columnwidth]{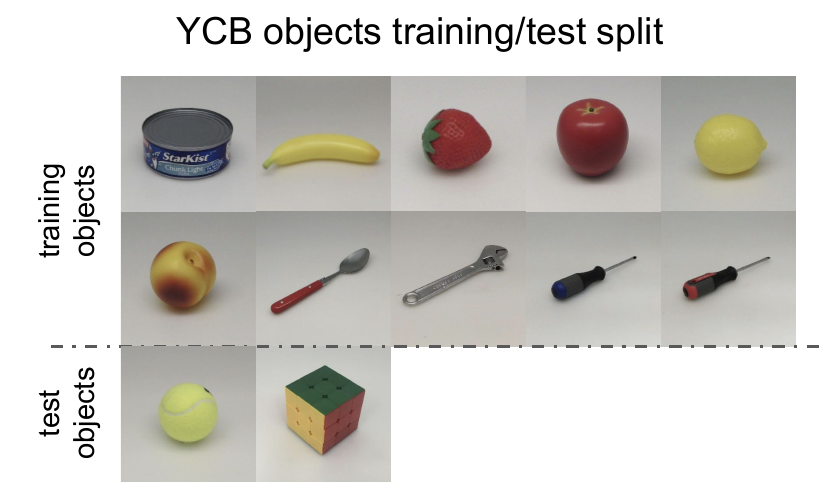}
	\caption{YCB objects photographs from the original list \url{www.ycbbenchmarks.com/object-set/}.
	The training objects are used as both distractor objects and target objects in our dataset.
	The test objects are used as target objects during evaluation.
	\label{fig:ycb_objects_split}}
\end{figure}

The embodiment we use is a Panda Franka Emika arm fitted with a Robotiq 2F-85 gripper.
The environment exposes a 7-DoF action space (which we discretize for prediction): 6-DoF Cartesian velocity control that is integrated into a reference pose that the end-effector tracks, and a 1-DoF velocity command to control the aperture of the parallel gripper.
The 6-DoF integrated velocity Cartesian controller builds on top of a 7-DoF integrated joint velocity controller.
The environment is equipped with 5 cameras, 2 on the front and one on the back of the basket, plus 2 cameras mounted on the gripper.
We used only two camera views in our experiments (see Figure~\ref{fig:ycb_episode}).

The YCB lifting task consists of lifting a specific object as specified in given instructions (e.g., "Lift the apple").
Multiple distractor objects (from 1 to 4) are present in the environment together with the target.
The distractor objects are randomly selected at every episode.
The 12 objects we use in the simulation are a subset of YCB dataset, in particular the food, kitchen and tool category.
See the training and test objects in Figure~\ref{fig:ycb_objects_split}.

The data has been generated via online RL using a variant of MPO~\citep{abdolmaleki2018maximum}.
The RL agent had access to camera views and proprioception information, including object poses.
The reward used is a staged reward, where the agent learns first to reach the object, then grasp it and finally lift it up to a certain height, while still keeping the object inside the basket.
This last condition was needed to avoid the agent learning to push the object on the basket slopes.
The resulting dataset includes 34,325 episodes, among which $94\%$ are successful.
The episodes have been generated for different target objects and using different numbers of distractors.

\subsection{Architecture}\label{sec:agent_exp_arch}

In all our agent experiments we use the multimodal agent architecture described in Section~\ref{sec:gats_agent_example} (and depicted in Figure~\ref{fig:gats_agent_example}), keep language and vision modules frozen, and train action and GATS modules from scratch.

\paragraph{Vision modules}
We explore the performance of GATS-based agents equipped with two distinct vision modules:
\begin{itemize}
    \item video-based agent utilizing the 1.1 billion parameter Phenaki model;
    \item image-based agent leveraging the 2.7 billion parameter ViT pretrained by us.
\end{itemize}
Since both vision modules have their own coupled language models (T5-like for Phenaki and Chinchilla for ViT) we reuse them for language processing.
See Section~\ref{sec:pretrained_models} for details about the vision foundation models.

\paragraph{Image processing}
We resize all visual observations from environments and datasets to an appropriate resolution that is compatible with the model we are using, i.e., 128x128 for Phenaki, and 256x256 for Parti tokenizer backing our ViT model.
Vision context length of the vision module is set to 5 frames when Phenaki is used, and 4 frames if ViT is used.

\paragraph{Atari adaptation}
Although Atari environments do not provide language instructions we keep using the same architecture and simply mask out the language tokens.

\paragraph{YCB adaptation}
The YCB environment provides multiple camera views and the agent architecture has to be adapted accordingly.
When using our GATS-based agent the necessary change is trivial.
We simply feed both camera views to the same vision module and expose obtained activations to GATS as separate modalities -- the steering mechanism seamlessly handles the rest.
While one could also relatively easily modify a traditional approach based on cross-attention by adding a new camera view, the processing of all visual inputs would be independent.
In contrast, GATS empowers each modality (including both camera views which are seen as separate modalities) with its own "voice" thanks to the symmetrical nature of the steering mechanism, allowing all modalities to dynamically share and refine information for a richer scene understanding.

\paragraph{Action head}
Action transformer outputs from the last layer are processed by a simple MLP to output logits for the actions.
We discretize action target values and utilize a standard cross-entropy loss for agent training.

\paragraph{Hyperparameters}
Hyperparameters and architecture details used to obtain the presented results are deferred to supplementary materials.

\subsection{Atari Pong results}

We begin our evaluation with the simple Atari Pong environment, a classic testbed for assessing the fundamentals of deep learning agents.
Here, we used an image-based agent that operates solely on visual and action modalities.
As mentioned in Section~\ref{sec:agent_exp_arch} language tokens are simply masked out and do not take part in the computations.

The agent quickly reaches human-expert performance, consistently achieving perfect scores.
This rapid mastery, exemplified in snapshots of a gameplay episodes in Figure~\ref{fig:atari_episode}, lays the groundwork for exploring GATS's capabilities in more complex environments where richer multi-modal interaction is required (presented in Sections~\ref{sec:agent_exp_lan_table}~and~\ref{sec:agent_exp_ycb}).

\subsection{Language-Table results}\label{sec:agent_exp_lan_table}

\begin{table}[ht!]
  \centering
  \caption{Language Table success rate (in \%).
  Average over total of 500 episodes.
  The results for two sets of ablation experiments are also presented.
  Agents with disabled vision steering perform significantly worse, unless a finetuned vision model is used.
  Classifier-free guidance leads to consistent improvements for both image and video based agents.
  \label{tab:lan_table}}
  \small
  \begin{tabular}{lccr}
	\toprule
	\multirow{2}{*}{Method} & Total & Unfrozen & \multirow{2}{*}{Reward} \\
	& params & params \\
	\midrule \midrule
	Image              & $3.1B$ & $495M$ & $89.0$ \\
	Video              & $1.3B$ & $182M$ & $76.8$ \\
	Video finetuned    & $1.3B$ & $182M$ & $75.6$ \\
	\midrule
	\multicolumn{4}{l}{\textit{Ablations: No vision steering}} \\
	\midrule
	Image              & $3.1B$ & $426M$ & $30.4$ \\
	Video              & $1.3B$ & $170M$ & $24.2$ \\
	Video finetuned    & $1.3B$ & $170M$ & $74.8$ \\
	\midrule
	\multicolumn{4}{l}{\textit{Ablations: No classifier-free guidance}} \\
	\midrule
	Image              & $3.1B$ & $495M$ & $83.2$ \\
	Video              & $1.3B$ & $182M$ & $73.6$ \\
	Video finetuned    & $1.3B$ & $182M$ & $75.0$ \\
	\bottomrule
  \end{tabular}
\end{table}

We conducted extensive experiments in the Language-Table environment, comparing the performance of GATS-based agents equipped with two distinct vision modules: ViT image model and Phenaki video model (see Section~\ref{sec:agent_exp_arch} for details).

Both GATS-based agents achieved state-of-the-art results.
The video-based agent reached a score of $76.8$, while the image-based agent achieved an even higher score of $89.0$ (see Table~\ref{tab:lan_table}).
In Figure~\ref{fig:lan_table_episode} we show three episodes handled by the image-based agent.
It is important to note that due to differences in pretraining datasets, model sizes, and even visual input resolution, these results should not be interpreted as suggesting a general superiority of image-based models over video-based models.
Rather, our use of distinct architectures highlights GATS's ability to effectively integrate diverse pretrained models.

Both agents achieved their results with the vision models remaining frozen, a feat enabled by a key innovation in GATS: the steering mechanism.
This mechanism alters the forward pass of pretrained models, granting the agent indirect access to their weights.
The advantages of steering become evident when comparing these results to performance of ablated agent where steering for the vision model is disabled (see Table~\ref{tab:lan_table}).
In such cases, the resulting architecture resembles a traditional cross-attention model, with the action model conditioned on features obtained from pretrained language and vision models.

Note that the Phenaki video model is a non-causal transformer pretrained with 11 input frames.
In our experiment, the input video sequence contains only the latest 5 frames and we also apply a temporal causal mask in order for efficient execution.
It is therefore not surprising that the GATS agent with the non-steered video component obtains low reward in this out-of-distribution downstream application.
In contrast, enabling steering allows GATS to improve the reward significantly without changing video model parameters at all.

To ablate the effect of the distribution mismatch, we run another experiment where we additionally finetuned the video model on Language-Table data without action labels.
This finetuning used the original MaskGIT~\citep{chang2022maskgit} loss with a causal mask to predict future frames of given episodes, essentially tailoring the visual features to the specific downstream task.
The finetuned video model is then frozen in the final GATS-based agent training.
The resulting agent without vision steering performs much better when leveraging the finetuned video model ($74.8$ vs $24.2$ score).
But it does not exceed the performance of the agent that steers the pretrained video model.
This finding reinforces the value of GATS, as it eliminates the need for additional finetuning and simplifies the integration of pretrained models.

To ensure comprehensiveness, we further evaluate a GATS-based agent that steers the finetuned video model.
Notably, this agent obtains a score of $75.6$ and hence does not achieve any improvement over the original agent steering the unaltered Phenaki video model.
This compelling result suggests that GATS's steering mechanism can effectively leverage pretrained models, even without the need for further task-specific fine-tuning.
This not only simplifies the training process but also underlines the generalizability of GATS across diverse visual representations.

Finally, we analyzed the impact of classifier-free guidance, observing consistent performance gains for both image and video agents, as illustrated in Table~\ref{tab:lan_table}.
Success rates in the Language Table environment increased by up to $5.8$ points, underlining the complementary nature of classifier-free guidance in ensuring goal-oriented action selection.
As mentioned in Section~\ref{sec:cfg} we set $\lambda$, the only hyperparameter for the method, to the value of $0.5$.

\begin{figure}[t]
	\centering
	\includegraphics[width=0.95\columnwidth]{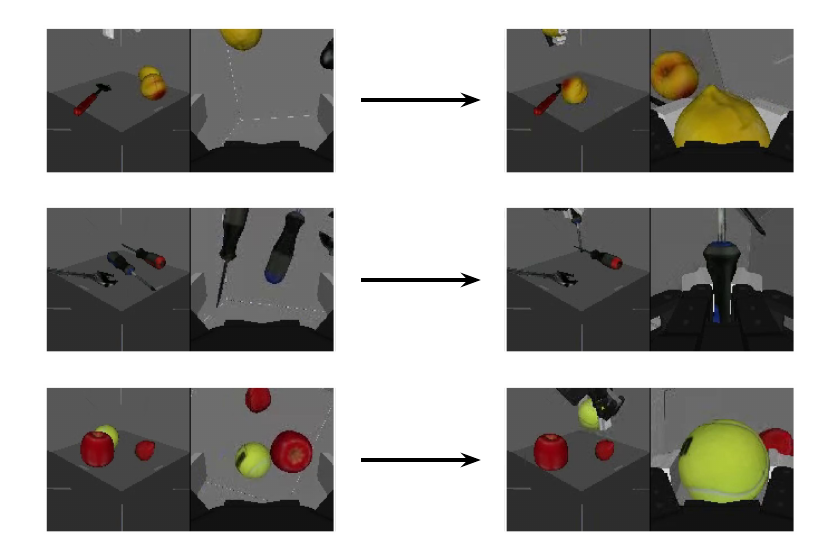}
	\caption{A GATS-based agent successfully lifts both training and test objects: a lemon, a screwdriver, and a tennis ball.
	On the left-hand side, there are two camera views from the beginning of a given episode, and on the right-hand size, there are the same camera views after the task completion.
	\label{fig:ycb_episode}}
\end{figure}

\subsection{YCB results}\label{sec:agent_exp_ycb}

Encouraged by successes of the GATS-based agent in the Language Table environment, we tackled the YCB environment, where the agent confronts a new challenge: orchestrating multiple camera views for manipulation of everyday objects.
Importantly, adapting our agent architecture to handle multiple cameras is extremely simple, as explained in Section~\ref{sec:agent_exp_arch}.

While quantitative reward metrics in YCB are context-dependent and not readily comparable across different research settings, we showcase the capabilities of our GATS video-based agent through examples of successful manipulation tasks (see Figure~\ref{fig:ycb_episode}).
As shown, the agent can skillfully lift and interact with a variety of objects, including lemons, screwdrivers, and tennis balls, demonstrating its ability to adapt to diverse shapes and textures.

YCB's emphasis on manipulation of everyday objects necessitates heightened precision and coordination.
Good results of our agent highlight the efficacy in this complex, multi-camera environment, paving the way for further exploration of its potential in real-world robotics applications.

\section{Image and caption generation}\label{sec:text_and_image}

\begin{figure}[hbt!]
	\centering
	\footnotesize
    \begin{minipage}{0.42\columnwidth}
	\includegraphics[width=\textwidth]{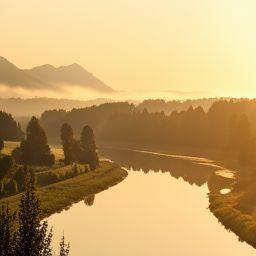}
	\end{minipage}%
	\hspace{0.03\columnwidth}%
	\begin{minipage}{0.55\columnwidth}
	\quad Entered textual prompt:\\
	\textit{Polish landscape with a mountain and a river bathed in the golden light of~sunrise.}
	\vskip 0.1cm
	\quad Generated caption:\\
	\textit{An image of a river and hills in \mbox{the morning.}}
	\end{minipage}
	
	\begin{minipage}{0.42\columnwidth}
	\includegraphics[width=\textwidth]{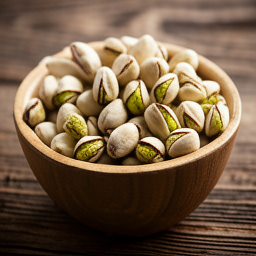}
	\end{minipage}%
	\hspace{0.03\columnwidth}%
	\begin{minipage}{0.55\columnwidth}
	\quad Entered textual prompt:\\
	\textit{Pistachios in a bowl on a rustic wooden table. Close-up commercial~shot.}
	\vskip 0.1cm
	\quad Generated caption:\\
	\textit{An image of some pistachios in a bowl, on wooden background.}
	\end{minipage}
	
	\begin{minipage}{0.42\columnwidth}
	\includegraphics[width=\textwidth]{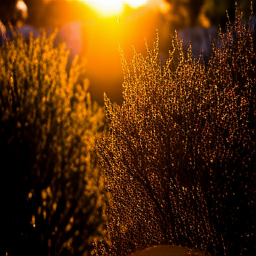}
	\end{minipage}%
	\hspace{0.03\columnwidth}%
	\begin{minipage}{0.55\columnwidth}
	\quad Entered textual prompt:\\
	\textit{A spotlight of sunset illuminates the delicate details of the~bush.}
	\vskip 0.1cm
	\quad Generated caption:\\
	\textit{An image of the morning sun \mbox{shining} through~a~bush.}
	\end{minipage}
	
	\begin{minipage}{0.42\columnwidth}
	\includegraphics[width=\textwidth]{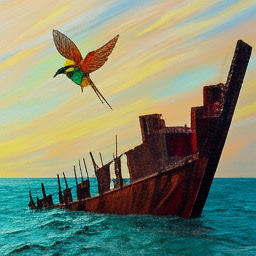}
	\end{minipage}%
	\hspace{0.03\columnwidth}%
	\begin{minipage}{0.55\columnwidth}
	\quad Entered textual prompt:\\
	\textit{A bee-eater hovering over a shipwreck. Oil painting in the style of~Rembrandt.}
	\vskip 0.1cm
	\quad Generated caption:\\
	\textit{An image of a flying bird in the sky over the decaying fishing~boat.}
	\end{minipage}
	
	\begin{minipage}{0.42\columnwidth}
	\includegraphics[width=\textwidth]{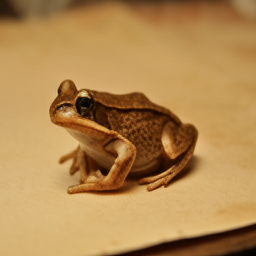}
	\end{minipage}%
	\hspace{0.03\columnwidth}%
	\begin{minipage}{0.55\columnwidth}
	\quad Entered textual prompt:\\
	\textit{A photo of a small frog sitting on an antique, yellowed piece \mbox{of paper}}.
	\vskip 0.1cm
	\quad Generated caption:\\
	\textit{An image of a small frog with a brown skin standing on a paper in a dark room, with a vintage~style.}
	\end{minipage}
	\caption{Five images generated by our 9.3B bimodal model using manually entered textual prompts (shown alongside each image).
	These images were subsequently used as visual prompts for the same model, with cleared context, to generate the captions presented alongside.
	The model was instructed to begin each caption with the \mbox{phrase "An image of"}.
	\label{fig:gen_image_caption}}
\end{figure}

Finally, to further demonstrate the versatility of GATS, we show how GATS can integrate text and image models to both process and generate either of these modalities.

We also describe the ViT foundation models that we pretrained, including the 2.7B parameter variant used in our agent experiments.

All hyperparameters and architecture details that are not included within the main text can be accessed in the supplementary materials.

\subsection{Bimodal vision-language model}\label{sec:text_image_bimodal}

The GATS-based bimodal model bridges the gap between image understanding and creation by fusing knowledge from two frozen foundation models: a 1.3B parameter Chinchilla language model and a 7.0B parameter ViT vision model.
Both foundation models share the same architecture, though the ViT model has a larger parameter count due to different hyperparameters.

The resulting bimodal model has 9.3 billion parameters, with only the GATS parameters (988M) being trained on the acquired image-text data, as both foundation models remain frozen.
We model language and vision jointly by processing each batch in two passes: first, with language tokens followed by vision tokens, and then vice versa.
Crucially, the same GATS layers are used in both passes.

Depending on the pass, we apply different training objectives.
When language tokens precede vision tokens, the MaskGIT loss~\citep{chang2022maskgit} guides the model to predict masked image tokens based on the provided caption and the visible image tokens.
Conversely, when vision tokens come first, a standard next token prediction cross-entropy loss aids the model to generate captions conditioned on the input image.
This approach allows the model to excel in both image captioning and text-to-image generation.

A key advantage of GATS lies in its symmetrical nature.
As mentioned before, GATS uses the same parameters for both image and text generation, facilitating efficient transfer learning between these tasks.

Figure~\ref{fig:gen_image_caption} showcases a selection of generated images.
They adhere closely to the given text prompts.
We used the generated images as visual prompts to the same bimodal model to produce the captions presented alongside in the same figure.
Additional captions produced using public domain images can be found in the supplementary materials.

\subsection{Pretrained vision transformers}

To train the foundation ViT models, we use the exact same architecture and data as in Section~\ref{sec:text_image_bimodal} but apply only the MaskGIT loss.
In this scenario, the language model remains frozen, while the vision model and the GATS parameters are trained from scratch.
Notably, the GATS layers are significantly smaller (124M parameters) when training for image generation alone, compared to the joint training setting (988M parameters).
This reflects the reduced complexity of image generation without the language modeling component.
We also train a smaller 2.7B parameter variant, with 1.3 billion parameters in the language model and another 1.3 billion dedicated to the trainable visual transformer.
This smaller variant is employed in the agent experiments discussed in Section~\ref{sec:agent_exp}.

\subsection{Extending GATS-based models}

While GATS can fuse any neural network, there are additional benefits when foundation models are GATS-based.
Namely, GATS parameters coupled with the pretrained modules can be substituted with a new single GATS model trained from scratch.
This approach not only simplifies processing but also centralizes all multimodal functionality within the new single GATS module.
This opens up new avenues for knowledge transfer across different multimodal tasks, as tasks like conditioning video on text may share inherent similarities with conditioning image on text.

\begin{figure}[t!]
	\centering
	\includegraphics[width=0.95\columnwidth]{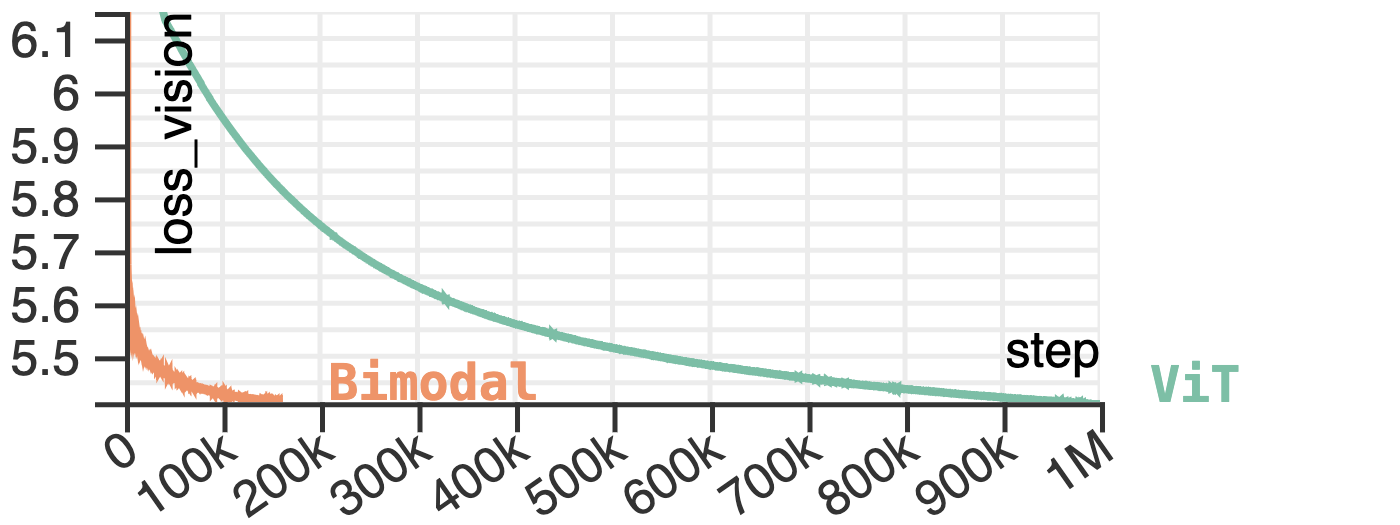}
	\vspace{0.01\columnwidth}
	\includegraphics[width=0.95\columnwidth]{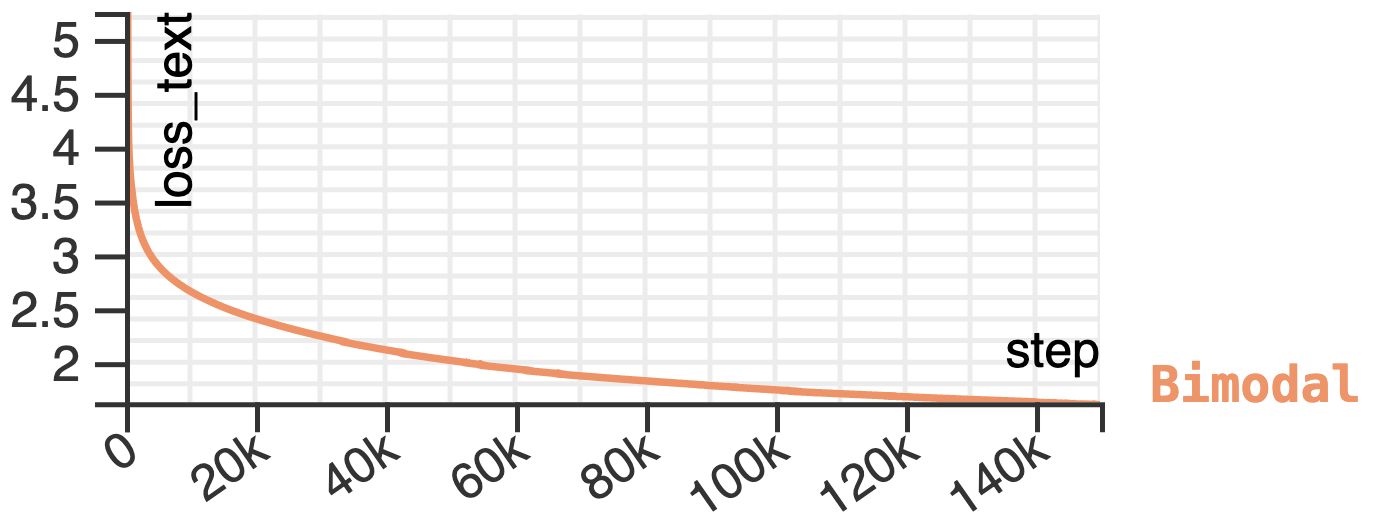}
	\caption{\textbf{Top.}~Our bimodal model (orange) rapidly matches ViT's (green) image generation performance.
    The MaskGIT loss for the bimodal model dips below 5.45 after only 50k steps, significantly faster than the 750k steps for ViT, in which the vision part is trained from scratch.
	This demonstrates that the single GATS module trained from scratch in the bimodal model can quickly recover the behavior of the substituted ViT's GATS.
	\textbf{Bottom.}~While achieving strong image generation, the bimodal model continues to refine its text generation capabilities.
    This highlights the model's ability to learn and improve both modalities simultaneously.
	\label{fig:image_and_caption_learning_curves}}
\end{figure}

We harness this benefit to construct the bimodal model described in Section~\ref{sec:text_image_bimodal}.
Instead of utilizing the entire 8.4B parameter ViT, we leverage the pretrained vision (7.0B) and language (1.3B) components, while discarding 124 million GATS parameters.
A new set of 988 million GATS parameters are then trained from scratch within this framework.
The resulting model swiftly recovers ViT's image generation performance while simultaneously acquiring and improving on its novel captioning ability, as demonstrated by the learning curves in Figure~\ref{fig:image_and_caption_learning_curves}.

\section{Related work}
\paragraph{Robotics with pretrained models}
There is a large body of literature on using pretrained representations for robotic control. 
The most common way of incorporating pretrained models is to use visual representations trained on general purpose data. 
CLIP \citep{radford2021learning} is a popular choice for various recent models \citep{shridhar2022cliport, khandelwal2022simple, gadre2022clip}. 
The Ego4D video dataset is also utilized as a source of both real world scenes and manipulation behavior \citep{nair2022r3m}.
RT-2 \citep{brohan2023rt} on the other hand, fine-tunes the weights of a pretrained vision-language model to be able to generate actions.
In contrast to the prior work, we demonstrate the use of multiple independently trained modalities, which are kept frozen within our novel architecture, and also show the effectiveness of pretrained generative video models for robotic control.

\paragraph{Adapter layers}
The adapter layer architecture we presented is primarily inspired by Flamingo \citep{Flamingo}.
Another example similar to our work is \citep{sharma2023lossless}, which uses adapter layers within vision networks to generate actions for robotics.
In our case, we propose a symmetric way to connect multiple modalities together.
In other words, all modalities can be generative models themselves and connected to generate a new modality as long as cross-modality data exists.

Another class of adapter models that is commonly used in the recent literature is LoRA \citep{hu2021lora}.
It adds parameters to the model as rank-decomposed feedforward operations in parallel to the pretrained layers.
In contrast to LoRA, GATS layers merge several different model layer outputs with an attention operation.
Our focus is the ability to use multiple modalities rather than parameter-efficient adaptation.

\paragraph{Foundation models}
Our work is built upon the tremendous progress in foundation models such as Chinchilla \citep{hoffmann2022training}, Flamingo \citep{Flamingo} and Phenaki \citep{villegas2022phenaki}.
We believe further development of both unimodal and multimodal generative models can pave the way for better agents across many domains, including robotics.

\section{Conclusion}

GATS is a novel and general-purpose method for combining pretrained foundation models.
It enables seamless integration of various modalities, such as language, vision, and action, without requiring any modification to the component models.
The steering mechanism in GATS allows for efficient coordination and knowledge transfer between modalities, resulting in enhanced performance in a variety of settings.

We have demonstrated the effectiveness of GATS through agent experiments in diverse environments: Atari Pong, Language-Table, and YCB.
We also show how GATS can integrate text and image models to process and generate any of these modalities.
Our results highlight the ability of GATS-based models to effectively leverage pretrained models and achieve state-of-the-art performance.

The modularity and extendibility of GATS-based architectures opens up exciting possibilities for future research.
It provides a flexible framework to incorporate additional modalities, leverage larger and more powerful foundation models, and explore novel applications that require the coordination and processing of multimodal inputs.

\section*{Acknowledgements}

We thank Zhishuai Zhang for his invaluable assistance in adapting the Parti VQ model, Ravi Addanki for conducting various explorations in the early stages of the project, and Giulia Vezzani for generously sharing her expertise and detailed information about the YCB environment and data.

Finally, we extend our appreciation to Nando de Freitas, Scott Reed, Jon Scholz and Vincent Vanhoucke for their thoughtful feedback on the final draft.
Their insights were greatly beneficial in refining our work and ensuring its clarity and~focus.

\section*{Author Contributions}

\textbf{Konrad~\.Zo\l{}na} developed the GATS architecture and implemented the initial prototype, planned and conducted the majority of experiments, and led the project overall.\\
\textbf{Serkan~Cabi} contributed to the design of the GATS architecture, trained and evaluated various action models, and contributed broadly to the codebase.\\
\textbf{Yutian~Chen} fine-tuned video models, implemented and conducted the initial GATS-based video agent experiments, helped in tuning and refining the GATS architecture, and contributed to paper writing.\\
\textbf{Eric~Lau} implemented the deterministic data pipeline across multiple TPU hosts that was used for training runs and prepared acquired image-text datasets for experiments.\\
\textbf{Claudio~Fantacci} curated the YCB robotics dataset and implemented parts of multi-host training and evaluation scripts, including logic for sharding models over multiple devices.\\
\textbf{Jurgis~Pasukonis} helped port the Phenaki video model for experiments.\\
\textbf{Jost~Tobias~Springenberg} implemented masked token prediction losses and parallel inference for image generation.\\
\textbf{Sergio~G\'{o}mez~Colmenarejo} was the technical lead of the project, heavily contributed to data, training, and evaluation infrastructure, and incorporated Language-Table datasets.

% Bibliography components
\bibliographystyle{abbrvnat}
\bibliography{refs}

\begin{thebibliography}{27}
\providecommand{\natexlab}[1]{#1}
\providecommand{\url}[1]{\texttt{#1}}
\expandafter\ifx\csname urlstyle\endcsname\relax
  \providecommand{\doi}[1]{doi: #1}\else
  \providecommand{\doi}{doi: \begingroup \urlstyle{rm}\Url}\fi

\bibitem[Abdolmaleki et~al.(2018)Abdolmaleki, Springenberg, Tassa, Munos,
  Heess, and Riedmiller]{abdolmaleki2018maximum}
A.~Abdolmaleki, J.~T. Springenberg, Y.~Tassa, R.~Munos, N.~Heess, and M.~A.
  Riedmiller.
\newblock Maximum a posteriori policy optimisation.
\newblock In \emph{ICLR}, 2018.

\bibitem[Alayrac et~al.(2022)Alayrac, Donahue, Luc, Miech, Barr, Hasson, Lenc,
  Mensch, Millican, Reynolds, Ring, Rutherford, Cabi, Han, Gong, Samangooei,
  Monteiro, Menick, Borgeaud, Brock, Nematzadeh, Sharifzadeh, Binkowski,
  Barreira, Vinyals, Zisserman, and Simonyan]{Flamingo}
J.~Alayrac, J.~Donahue, P.~Luc, A.~Miech, I.~Barr, Y.~Hasson, K.~Lenc,
  A.~Mensch, K.~Millican, M.~Reynolds, R.~Ring, E.~Rutherford, S.~Cabi, T.~Han,
  Z.~Gong, S.~Samangooei, M.~Monteiro, J.~L. Menick, S.~Borgeaud, A.~Brock,
  A.~Nematzadeh, S.~Sharifzadeh, M.~Binkowski, R.~Barreira, O.~Vinyals,
  A.~Zisserman, and K.~Simonyan.
\newblock Flamingo: a visual language model for few-shot learning.
\newblock In \emph{NeurIPS}, 2022.

\bibitem[Bellemare et~al.(2013)Bellemare, Naddaf, Veness, and
  Bowling]{bellemare2013arcade}
M.~G. Bellemare, Y.~Naddaf, J.~Veness, and M.~Bowling.
\newblock The arcade learning environment: An evaluation platform for general
  agents.
\newblock \emph{J. Artif. Intell. Res.}, 2013.

\bibitem[Brohan et~al.(2023)Brohan, Brown, Carbajal, Chebotar, Chen,
  Choromanski, Ding, Driess, Dubey, Finn, Florence, Fu, Arenas, Gopalakrishnan,
  Han, Hausman, Herzog, Hsu, Ichter, Irpan, Joshi, Julian, Kalashnikov, Kuang,
  Leal, Lee, Lee, Levine, Lu, Michalewski, Mordatch, Pertsch, Rao, Reymann,
  Ryoo, Salazar, Sanketi, Sermanet, Singh, Singh, Soricut, Tran, Vanhoucke,
  Vuong, Wahid, Welker, Wohlhart, Wu, Xia, Xiao, Xu, Xu, Yu, and
  Zitkovich]{brohan2023rt}
A.~Brohan, N.~Brown, J.~Carbajal, Y.~Chebotar, X.~Chen, K.~Choromanski,
  T.~Ding, D.~Driess, A.~Dubey, C.~Finn, P.~Florence, C.~Fu, M.~G. Arenas,
  K.~Gopalakrishnan, K.~Han, K.~Hausman, A.~Herzog, J.~Hsu, B.~Ichter,
  A.~Irpan, N.~J. Joshi, R.~Julian, D.~Kalashnikov, Y.~Kuang, I.~Leal, L.~Lee,
  T.~E. Lee, S.~Levine, Y.~Lu, H.~Michalewski, I.~Mordatch, K.~Pertsch, K.~Rao,
  K.~Reymann, M.~S. Ryoo, G.~Salazar, P.~Sanketi, P.~Sermanet, J.~Singh,
  A.~Singh, R.~Soricut, H.~T. Tran, V.~Vanhoucke, Q.~Vuong, A.~Wahid,
  S.~Welker, P.~Wohlhart, J.~Wu, F.~Xia, T.~Xiao, P.~Xu, S.~Xu, T.~Yu, and
  B.~Zitkovich.
\newblock {RT-2:} vision-language-action models transfer web knowledge to
  robotic control.
\newblock \emph{arXiv, 2307.15818}, 2023.

\bibitem[Chang et~al.(2022)Chang, Zhang, Jiang, Liu, and
  Freeman]{chang2022maskgit}
H.~Chang, H.~Zhang, L.~Jiang, C.~Liu, and W.~T. Freeman.
\newblock Maskgit: Masked generative image transformer.
\newblock In \emph{CVPR}, 2022.

\bibitem[Dosovitskiy et~al.(2021)Dosovitskiy, Beyer, Kolesnikov, Weissenborn,
  Zhai, Unterthiner, Dehghani, Minderer, Heigold, Gelly, Uszkoreit, and
  Houlsby]{dosovitskiy2021image}
A.~Dosovitskiy, L.~Beyer, A.~Kolesnikov, D.~Weissenborn, X.~Zhai,
  T.~Unterthiner, M.~Dehghani, M.~Minderer, G.~Heigold, S.~Gelly, J.~Uszkoreit,
  and N.~Houlsby.
\newblock An image is worth 16x16 words: Transformers for image recognition at
  scale.
\newblock In \emph{ICLR}, 2021.

\bibitem[Gadre et~al.(2022)Gadre, Wortsman, Ilharco, Schmidt, and
  Song]{gadre2022clip}
S.~Y. Gadre, M.~Wortsman, G.~Ilharco, L.~Schmidt, and S.~Song.
\newblock {CLIP} on wheels: Zero-shot object navigation as object localization
  and exploration.
\newblock \emph{arXiv, 2203.10421}, 2022.

\bibitem[Hendrycks and Gimpel(2016)]{hendrycks2023gaussian}
D.~Hendrycks and K.~Gimpel.
\newblock Bridging nonlinearities and stochastic regularizers with gaussian
  error linear units.
\newblock \emph{arXiv, 1606.08415}, 2016.

\bibitem[Hessel et~al.(2021)Hessel, Danihelka, Viola, Guez, Schmitt, Sifre,
  Weber, Silver, and van Hasselt]{hessel2021muesli}
M.~Hessel, I.~Danihelka, F.~Viola, A.~Guez, S.~Schmitt, L.~Sifre, T.~Weber,
  D.~Silver, and H.~van Hasselt.
\newblock Muesli: Combining improvements in policy optimization.
\newblock In \emph{ICML}, 2021.

\bibitem[Ho and Salimans(2022)]{ho2022classifier}
J.~Ho and T.~Salimans.
\newblock Classifier-free diffusion guidance.
\newblock \emph{arXiv, 2207.12598}, 2022.

\bibitem[Hoffmann et~al.(2022)Hoffmann, Borgeaud, Mensch, Buchatskaya, Cai,
  Rutherford, de~Las~Casas, Hendricks, Welbl, Clark, Hennigan, Noland,
  Millican, van~den Driessche, Damoc, Guy, Osindero, Simonyan, Elsen, Rae,
  Vinyals, and Sifre]{hoffmann2022training}
J.~Hoffmann, S.~Borgeaud, A.~Mensch, E.~Buchatskaya, T.~Cai, E.~Rutherford,
  D.~de~Las~Casas, L.~A. Hendricks, J.~Welbl, A.~Clark, T.~Hennigan, E.~Noland,
  K.~Millican, G.~van~den Driessche, B.~Damoc, A.~Guy, S.~Osindero,
  K.~Simonyan, E.~Elsen, J.~W. Rae, O.~Vinyals, and L.~Sifre.
\newblock Training compute-optimal large language models.
\newblock \emph{arXiv, 2203.15556}, 2022.

\bibitem[Hu et~al.(2022)Hu, Shen, Wallis, Allen{-}Zhu, Li, Wang, Wang, and
  Chen]{hu2021lora}
E.~J. Hu, Y.~Shen, P.~Wallis, Z.~Allen{-}Zhu, Y.~Li, S.~Wang, L.~Wang, and
  W.~Chen.
\newblock Lora: Low-rank adaptation of large language models.
\newblock In \emph{ICLR}, 2022.

\bibitem[Khandelwal et~al.(2022)Khandelwal, Weihs, Mottaghi, and
  Kembhavi]{khandelwal2022simple}
A.~Khandelwal, L.~Weihs, R.~Mottaghi, and A.~Kembhavi.
\newblock Simple but effective: {CLIP} embeddings for embodied {AI}.
\newblock In \emph{CVPR}, 2022.

\bibitem[Kingma and Ba(2015)]{kingma2017adam}
D.~P. Kingma and J.~Ba.
\newblock Adam: {A} method for stochastic optimization.
\newblock In \emph{ICLR}, 2015.

\bibitem[Lifshitz et~al.(2023)Lifshitz, Paster, Chan, Ba, and
  McIlraith]{lifshitz2023steve}
S.~Lifshitz, K.~Paster, H.~Chan, J.~Ba, and S.~A. McIlraith.
\newblock {STEVE-1:} {A} generative model for text-to-behavior in minecraft.
\newblock \emph{arXiv, 2306.00937}, 2023.

\bibitem[Lynch et~al.(2022)Lynch, Wahid, Tompson, Ding, Betker, Baruch,
  Armstrong, and Florence]{lynch2022interactive}
C.~Lynch, A.~Wahid, J.~Tompson, T.~Ding, J.~Betker, R.~Baruch, T.~Armstrong,
  and P.~Florence.
\newblock Interactive language: Talking to robots in real time.
\newblock \emph{arXiv, 2210.06407}, 2022.

\bibitem[Nair et~al.(2022)Nair, Rajeswaran, Kumar, Finn, and
  Gupta]{nair2022r3m}
S.~Nair, A.~Rajeswaran, V.~Kumar, C.~Finn, and A.~Gupta.
\newblock {R3M:} {A} universal visual representation for robot manipulation.
\newblock In \emph{CoRL}, 2022.

\bibitem[Radford et~al.(2021)Radford, Kim, Hallacy, Ramesh, Goh, Agarwal,
  Sastry, Askell, Mishkin, Clark, Krueger, and Sutskever]{radford2021learning}
A.~Radford, J.~W. Kim, C.~Hallacy, A.~Ramesh, G.~Goh, S.~Agarwal, G.~Sastry,
  A.~Askell, P.~Mishkin, J.~Clark, G.~Krueger, and I.~Sutskever.
\newblock Learning transferable visual models from natural language
  supervision.
\newblock In \emph{ICML}, 2021.

\bibitem[Raffel et~al.(2020)Raffel, Shazeer, Roberts, Lee, Narang, Matena,
  Zhou, Li, and Liu]{raffel2020exploring}
C.~Raffel, N.~Shazeer, A.~Roberts, K.~Lee, S.~Narang, M.~Matena, Y.~Zhou,
  W.~Li, and P.~J. Liu.
\newblock Exploring the limits of transfer learning with a unified text-to-text
  transformer.
\newblock \emph{J. Mach. Learn. Res.}, 2020.

\bibitem[Reed et~al.(2022)Reed, Zolna, Parisotto, Colmenarejo, Novikov,
  Barth{-}Maron, Gimenez, Sulsky, Kay, Springenberg, Eccles, Bruce, Razavi,
  Edwards, Heess, Chen, Hadsell, Vinyals, Bordbar, and de~Freitas]{reed2022}
S.~E. Reed, K.~Zolna, E.~Parisotto, S.~G. Colmenarejo, A.~Novikov,
  G.~Barth{-}Maron, M.~Gimenez, Y.~Sulsky, J.~Kay, J.~T. Springenberg,
  T.~Eccles, J.~Bruce, A.~Razavi, A.~Edwards, N.~Heess, Y.~Chen, R.~Hadsell,
  O.~Vinyals, M.~Bordbar, and N.~de~Freitas.
\newblock A generalist agent.
\newblock \emph{Trans. Mach. Learn. Res.}, 2022.

\bibitem[Sharma et~al.(2023)Sharma, Fantacci, Zhou, Koppula, Heess, Scholz, and
  Aytar]{sharma2023lossless}
M.~Sharma, C.~Fantacci, Y.~Zhou, S.~Koppula, N.~Heess, J.~Scholz, and Y.~Aytar.
\newblock Lossless adaptation of pretrained vision models for robotic
  manipulation.
\newblock In \emph{ICLR}, 2023.

\bibitem[Shridhar et~al.(2021)Shridhar, Manuelli, and Fox]{shridhar2022cliport}
M.~Shridhar, L.~Manuelli, and D.~Fox.
\newblock Cliport: What and where pathways for robotic manipulation.
\newblock In \emph{CoRL}, 2021.

\bibitem[Todorov et~al.(2012)Todorov, Erez, and Tassa]{todorov2012mujoco}
E.~Todorov, T.~Erez, and Y.~Tassa.
\newblock Mujoco: {A} physics engine for model-based control.
\newblock In \emph{IROS}, 2012.

\bibitem[Vaswani et~al.(2017)Vaswani, Shazeer, Parmar, Uszkoreit, Jones, Gomez,
  Kaiser, and Polosukhin]{vaswani2017attention}
A.~Vaswani, N.~Shazeer, N.~Parmar, J.~Uszkoreit, L.~Jones, A.~N. Gomez,
  L.~Kaiser, and I.~Polosukhin.
\newblock Attention is all you need.
\newblock In \emph{NeurIPS}, 2017.

\bibitem[Villegas et~al.(2023)Villegas, Babaeizadeh, Kindermans, Moraldo,
  Zhang, Saffar, Castro, Kunze, and Erhan]{villegas2022phenaki}
R.~Villegas, M.~Babaeizadeh, P.~Kindermans, H.~Moraldo, H.~Zhang, M.~T. Saffar,
  S.~Castro, J.~Kunze, and D.~Erhan.
\newblock Phenaki: Variable length video generation from open domain textual
  descriptions.
\newblock In \emph{ICLR}, 2023.

\bibitem[Xiong et~al.(2020)Xiong, Yang, He, Zheng, Zheng, Xing, Zhang, Lan,
  Wang, and Liu]{xiong2020layer}
R.~Xiong, Y.~Yang, D.~He, K.~Zheng, S.~Zheng, C.~Xing, H.~Zhang, Y.~Lan,
  L.~Wang, and T.~Liu.
\newblock On layer normalization in the transformer architecture.
\newblock In \emph{ICML}, 2020.

\bibitem[Yu et~al.(2022)Yu, Xu, Koh, Luong, Baid, Wang, Vasudevan, Ku, Yang,
  Ayan, Hutchinson, Han, Parekh, Li, Zhang, Baldridge, and Wu]{yu2022scaling}
J.~Yu, Y.~Xu, J.~Y. Koh, T.~Luong, G.~Baid, Z.~Wang, V.~Vasudevan, A.~Ku,
  Y.~Yang, B.~K. Ayan, B.~Hutchinson, W.~Han, Z.~Parekh, X.~Li, H.~Zhang,
  J.~Baldridge, and Y.~Wu.
\newblock Scaling autoregressive models for content-rich text-to-image
  generation.
\newblock \emph{Trans. Mach. Learn. Res.}, 2022.

\end{thebibliography}

\clearpage
\section*{Supplementary Materials}

\subsection*{Hyperparameter and architecture details}

This section provides detailed hyperparameter and architecture specifications.
We focus on the models that we trained directly, and the details for pretrained models like Chinchilla and Phenaki can be found in their respective papers.
All models rely on transformers~\citep{vaswani2017attention} with pre-layer norms as proposed by \citet{xiong2020layer}, GELU activations~\citep{hendrycks2023gaussian}, and shared encoder-decoder embeddings.
We used the Adam optimizer~\citep{kingma2017adam} for training across all models.

\paragraph{ViT}
We pre-train our vision transformers on acquired image-text data using a batch size of $2048$ and a learning rate of $1\mathrm{e}{-5}$ for 1 million steps.
We train two model variants with different sizes: \texttt{ViT-2.7B} has 2.7 billion parameters and \texttt{ViT-8.5B} has 8.5 billion parameters.

The detailed architecture specifications of the vision modules are presented in Table~\ref{tab:arch-ViT-vision}.
Both ViT models use the same GATS hyperparameters during vision pre-training.
However, a larger GATS module is used for the 9.4B bimodal model presented in Section~\ref{sec:text_and_image}.
Table~\ref{tab:arch-ViT-gats} compares the hyperparameters of the GATS transformers used in different settings.

\paragraph{Agents}
The architecture hyperparameters of the action and GATS modules in our agent experiments are tailored to the type of vision module used (image-based or video-based).
While these hyperparameters differ between vision modules, they remain consistent across all experiments.

Table~\ref{tab:arch-agent-action} outlines the architecture details of the action modules.
As we do not use proprioception values in the tasks considered, we input a fixed number of trainable input embeddings for each time step: 8 for image-based agents and 16 for video-based agents.
We do not input past actions and hence aforementioned input embeddings are the only action inputs.
Notably, the video action submodule is factorized across time and space to conserve device memory.
This factorization involves applying attention masks such that even layers attend only over space (tokens within the same frame) while odd layers attend only over time (tokens across different time steps for the same location).
The architecture specifications of the GATS modules are presented in Table~\ref{tab:arch-agent-gats}.
\begin{table}[t]
    \centering
    \centering
	\caption{ViT vision module hyperparameters in \texttt{ViT-8.5B} and \texttt{ViT-2.7B}.
	\label{tab:arch-ViT-vision}}\vskip -0.1cm
	\small
    \begin{tabular}{l|cc}
    \toprule
    \textbf{Hyperparameter} & \texttt{ViT-8.5B} & \texttt{ViT-2.7B} \\
    \midrule
    Transformer blocks  & 32 & 18 \\
    Attention heads     & 32 & 18 \\
    Layer width         & 4 096 & 2 304 \\
    MLP hidden size     & 24 576 & 13 824 \\
    \bottomrule
    \end{tabular}\vskip 0.2cm
	\caption{ViT GATS module hyperparameters.
	\label{tab:arch-ViT-gats}}\vskip 0.1cm
    \begin{tabular}{l|cc}
    \toprule
    \multirow{2}{*}{\textbf{Hyperparameter}} & Vision & Bimodal \\
    & pretraining & finetuning \\
    \midrule
    Transformer blocks  & 12 & 12 \\
    Attention heads     & 8 & 16 \\
    Layer width         & 512 & 2 048 \\
    MLP hidden size     & 3 072 & 12 288 \\
    \bottomrule
    \end{tabular}\vskip 0.2cm
	\caption{Action module hyperparameters.
	\label{tab:arch-agent-action}}\vskip 0.1cm
	\small
    \begin{tabular}{l|cc}
    \toprule
    \textbf{Hyperparameter} & Image & Video \\
    \midrule
    Transformer blocks  & 24 & 24 \\
    Attention heads     & 32 & 4 \\
    Layer width         & 1 024 & 512 \\
    MLP hidden size     & 6 144 & 3 072 \\
    Time-space factorization & No & Yes \\
    \bottomrule
    \end{tabular}\vskip 0.2cm
	\caption{Agent GATS module hyperparameters.
	\label{tab:arch-agent-gats}}\vskip 0.1cm
    \begin{tabular}{l|cc}
    \toprule
    \textbf{Hyperparameter} & Image & Video \\
    \midrule
    Transformer blocks  & 18 & 12 \\
    Attention heads     & 8 & 8 \\
    Layer width         & 512 & 512 \\
    MLP hidden size     & 3 072 & 3 072 \\
    \bottomrule
    \end{tabular}
\end{table}

Both agent types are trained with a learning rate of 1e-4, including a linear warm-up of 750 steps.
However, the image-based agent is trained for 100k steps, while the video-based agent is trained for 250k steps.
Batch sizes also vary: 1024 for the image-based agent (on both Atari and Language Table), 512 for the video-based agent on Language Table, and 256 for the video-based agent on YCB (due to increased memory demands from two camera streams).

\subsection*{Public Domain Image Captioning}

To evaluate the captioning capabilities of our 9.3B bimodal model (introduced in Section~\ref{sec:text_image_bimodal}), we present captions generated for images sourced from Wikimedia Commons, the free media repository.
These images were featured as Pictures of the Day\footnote{See all archived Pictures of the Day on Wikimedia Commons: \url{commons.wikimedia.org/wiki/Commons:Picture_of_the_day}} between January 6th and January 10th, 2024.
We verified that these images were not present in the training~dataset.

\begin{figure}[t]
	\centering
	\footnotesize
    \begin{minipage}{0.42\columnwidth}
	\includegraphics[width=\textwidth,height=\textwidth]{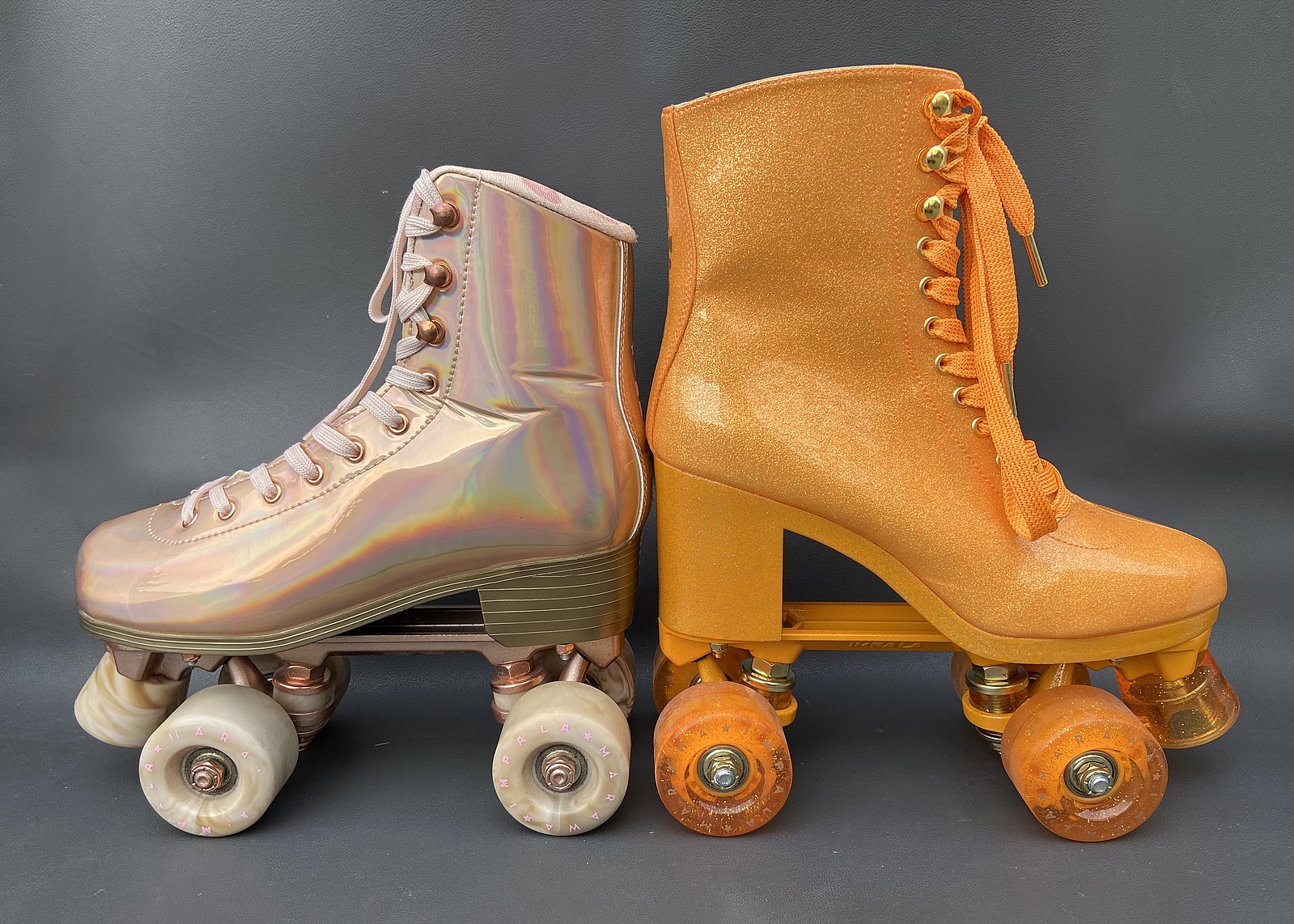}
	\end{minipage}%
	\hspace{0.03\columnwidth}%
	\begin{minipage}{0.55\columnwidth}
	\quad Original description:\\
	\textit{A rose gold roller skate compared with an orange high-heeled~one.}
	\vskip 0.1cm
	\quad Generated caption:\\
	\textit{An image of roller skates on a gray background.}
	\end{minipage}
	
	\begin{minipage}{0.42\columnwidth}
	\includegraphics[width=\textwidth,height=\textwidth]{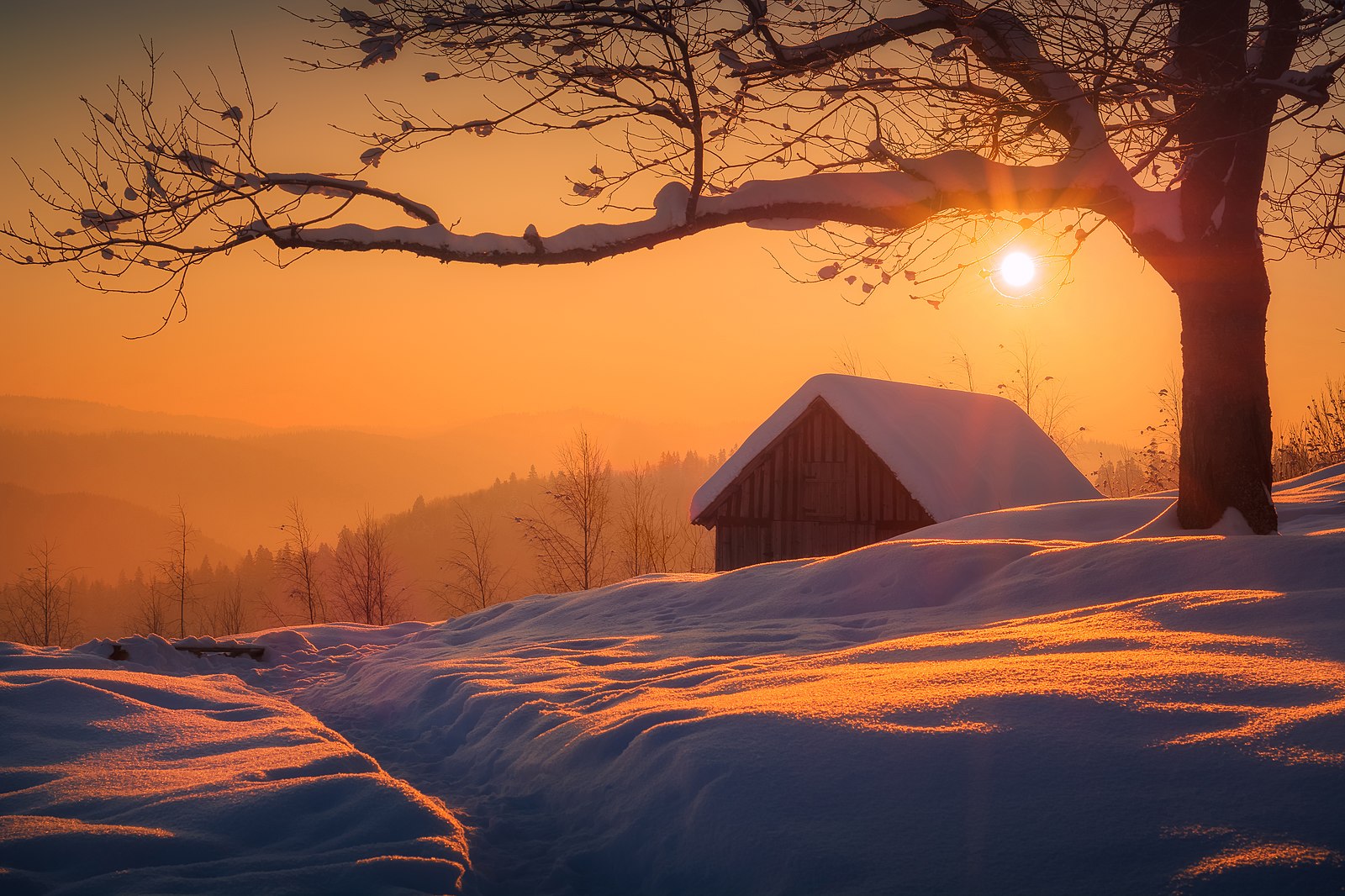}
	\end{minipage}%
	\hspace{0.03\columnwidth}%
	\begin{minipage}{0.55\columnwidth}
	\quad Original description:\\
	\textit{Winter morning in the Carpathian National Park, Ivano-Frankivsk Oblast,~Ukraine.}
	\vskip 0.1cm
	\quad Generated caption:\\
	\textit{An image of a nice winter scenery against morning sun light. Nice warm colors, beautiful scenery.}
	\end{minipage}
	
	\begin{minipage}{0.42\columnwidth}
	\includegraphics[width=\textwidth,height=\textwidth]{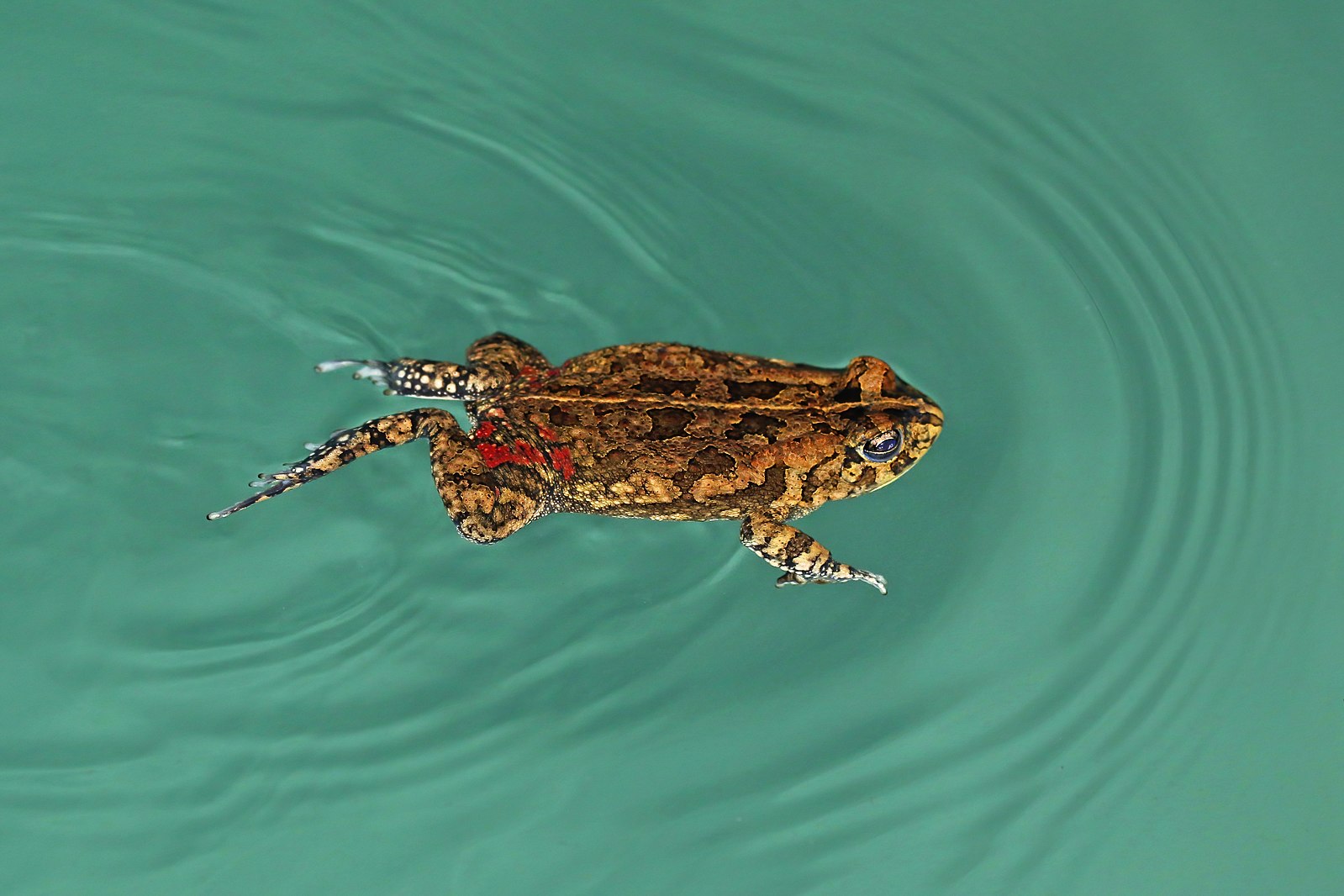}
	\end{minipage}%
	\hspace{0.03\columnwidth}%
	\begin{minipage}{0.55\columnwidth}
	\quad Original description:\\
	\textit{An African common toad swimming in Lake Sibhayi, South~Africa.}
	\vskip 0.1cm
	\quad Generated caption:\\
	\textit{An image of a frog is taken with a high angled view.}
	\end{minipage}
	
	\begin{minipage}{0.42\columnwidth}
	\includegraphics[width=\textwidth,height=\textwidth]{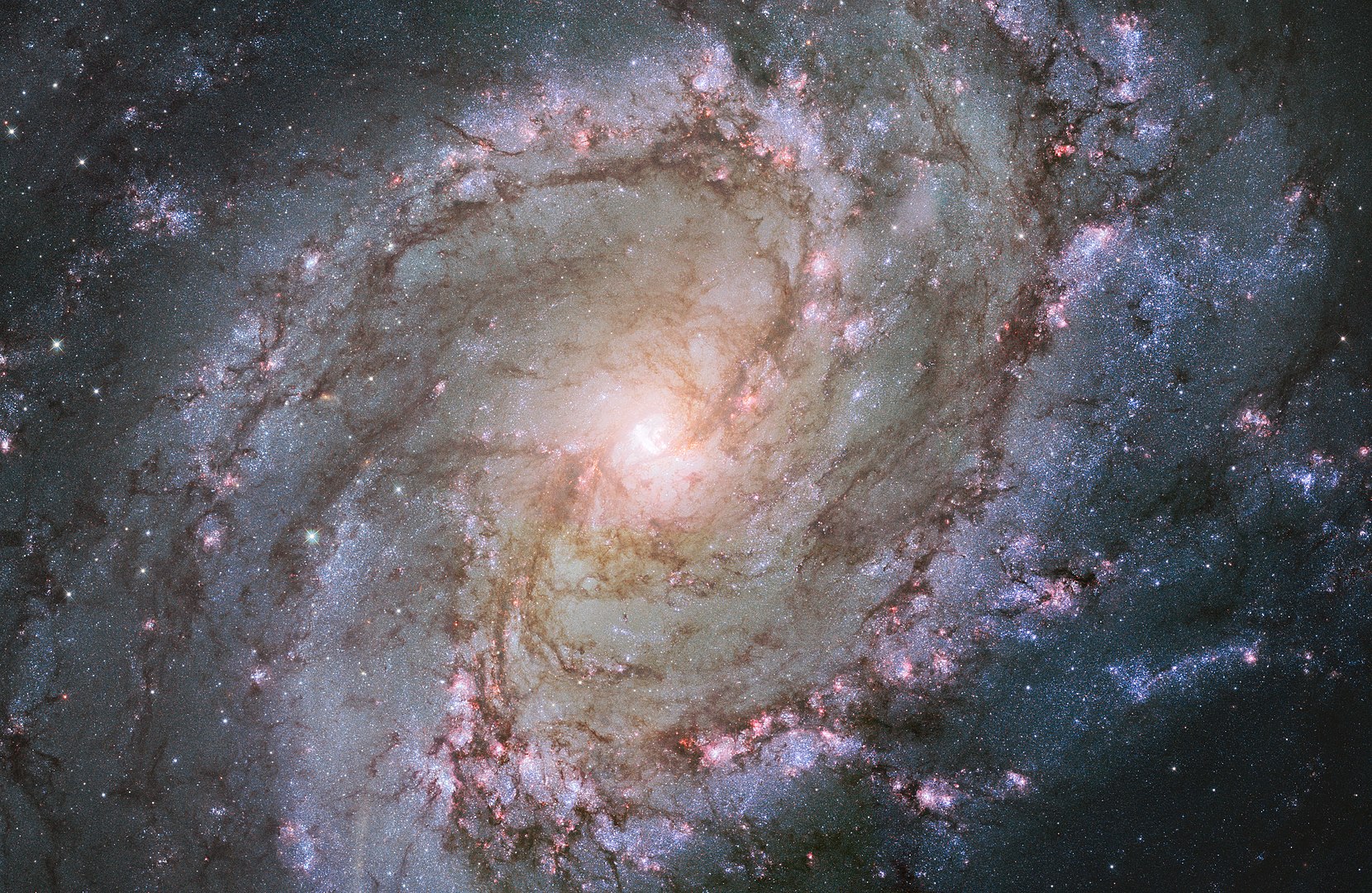}
	\end{minipage}%
	\hspace{0.03\columnwidth}%
	\begin{minipage}{0.55\columnwidth}
	\quad Original description:\\
	\textit{This new Hubble image shows the scatterings of bright stars and thick dust that make up spiral galaxy~Messier~83.}
	\vskip 0.1cm
	\quad Generated caption:\\
	\textit{An image of a spiral galaxy in \mbox{the universe.}}
	\end{minipage}
	
	\begin{minipage}{0.42\columnwidth}
	\includegraphics[width=\textwidth,height=\textwidth]{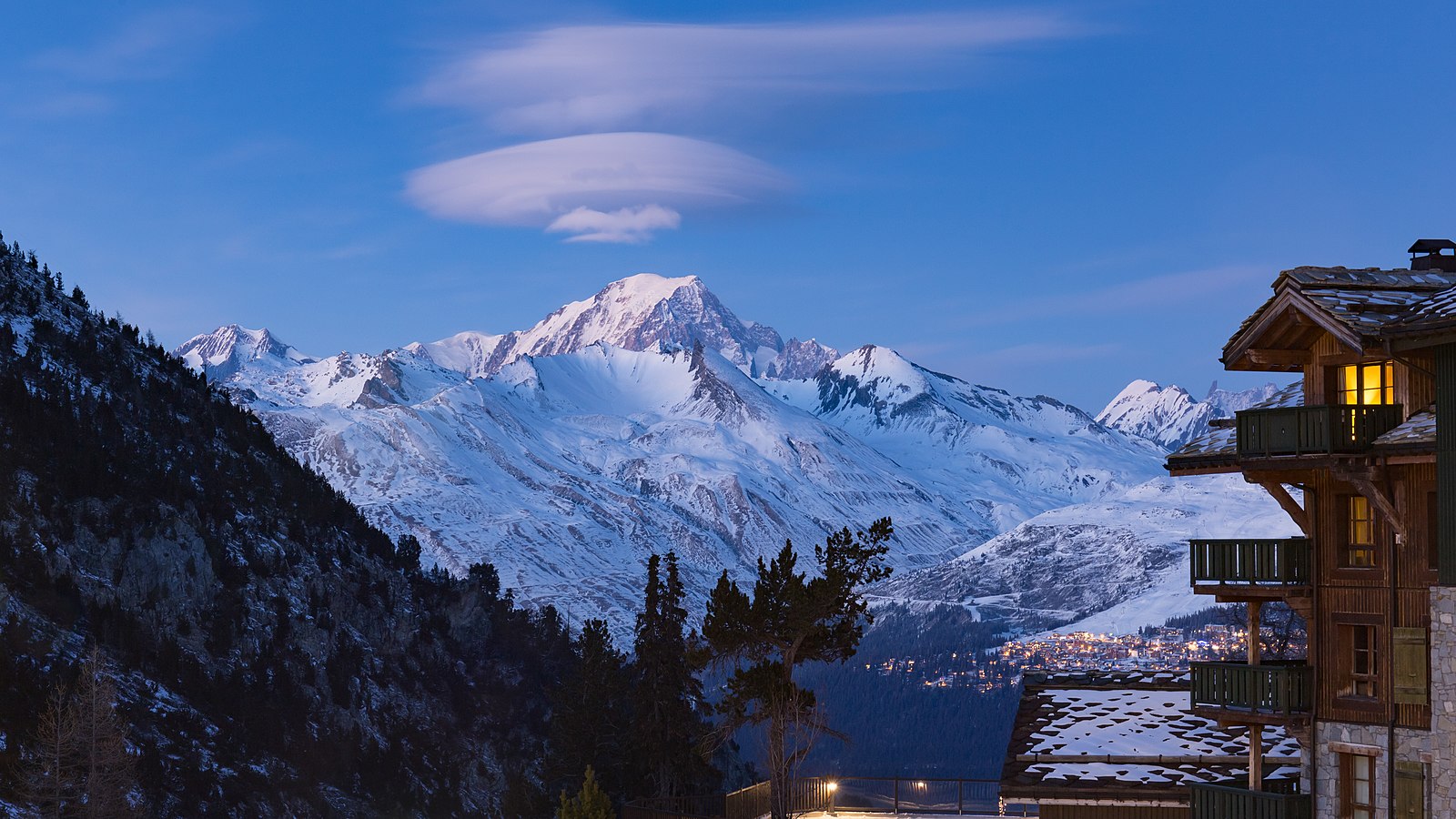}
	\end{minipage}%
	\hspace{0.03\columnwidth}%
	\begin{minipage}{0.55\columnwidth}
	\quad Original description:\\
	\textit{The Mont Blanc ("white mountain") located on the French-Italian~border.}
	\vskip 0.1cm
	\quad Generated caption:\\
	\textit{An image of mountain and a cloud on the sky at morning. Alpine ski~resort.}
	\end{minipage}
	\caption{Captions produced by our 9.3B bimodal model for five images sourced from Wikimedia Commons.
    Generated captions are presented alongside their corresponding prompt images and original English descriptions.
    The model was instructed to begin each caption with the \mbox{phrase "An image of"}.
    All images were resized to the square format required by our~model.
	\label{fig:gen_pub_dom_image_caption}}
	\vskip 2cm
\end{figure}

Figure~\ref{fig:gen_pub_dom_image_caption} displays the generated captions alongside their original English descriptions.
The generated captions accurately reflect the visual elements present in the images, highlighting our model's proficiency in image understanding and natural language generation.

\end{document}